\documentclass[10pt,twocolumn,letterpaper]{article}

\usepackage{iccv}
\usepackage{times}
\usepackage{epsfig}
\usepackage{graphicx}
\usepackage{amsmath}
\usepackage{amssymb}
\usepackage{multirow}
\usepackage{physics}
\usepackage{nicefrac}
\usepackage{algorithm}
\usepackage{xcolor}
\usepackage{pifont}
\usepackage{tikz}
\usepackage{subcaption}
\usepackage{tabularx}
\usepackage{comment}
\usepackage{graphicx}
\usepackage{makecell}
\usepackage[export]{adjustbox}

\usepackage{color}
\usepackage[normalem]{ulem}
\usepackage{multibib}
\usepackage{graphbox}
\usepackage{paralist}
\usepackage{setspace}
\usepackage[noend]{algpseudocode}
\usepackage{xpatch}
\usepackage{enumitem}

\usepackage{colortbl}
\usepackage{arydshln,graphicx,xcolor,array}
\usetikzlibrary{calc, fit}

\usepackage[pagebackref,breaklinks,colorlinks]{hyperref}
\usepackage[capitalize]{cleveref}
\crefname{section}{Sec.}{Secs.}
\Crefname{section}{Section}{Sections}
\Crefname{table}{Table}{Tables}
\crefname{table}{Tab.}{Tabs.}

\definecolor{MyGreen}{cmyk}{100, 0, 100, 0}
\definecolor{DarkGreen}{cmyk}{100, 0, 10, 0}

\definecolor{darkergreen}{RGB}{21, 152, 56}
\definecolor{red2}{RGB}{252, 54, 65}
\definecolor{right_hand_blue}{RGB}{44, 126, 183}
\definecolor{simtoken_red}{RGB}{213, 40, 38}
\definecolor{jointoken_green}{RGB}{82, 173, 60}
\definecolor{left_hand_orange}{RGB}{255, 130, 22}
\definecolor{shared_hand_yellow}{RGB}{247,202,23}
\newcommand{\cmark}{\textcolor{darkergreen}{\ding{51}}}%
\newcommand{\xmark}{\textcolor{red2}{\ding{55}}}%

\iccvfinalcopy 

\def\iccvPaperID{8} 

\ificcvfinal\fi

\begin{document}

\title{Extract-and-Adaptation Network for 3D Interacting Hand Mesh Recovery}

\author{
  JoonKyu Park$^{1*}$ \hskip1.6em Daniel Sungho Jung$^{2,3*}$ \hskip1.6em Gyeongsik Moon$^{4*}$ \hskip1.6em Kyoung Mu Lee$^{1,2,3}$ \\
   $^{1}$Dept. of ECE\&ASRI, $^{2}$IPAI, Seoul National University, Korea \\ 
   $^{3}$SNU-LG AI Research Center, 
   $^{4}$Meta Reality Labs Research \\
   {\tt\small \{jkpark0825, dqj5182\}@snu.ac.kr, mks0601@meta.com, kyoungmu@snu.ac.kr} 
}

\maketitle
\ificcvfinal\thispagestyle{empty}\fi

\begin{abstract}
Understanding how two hands interact with each other is a key component of accurate 3D interacting hand mesh recovery.
However, recent Transformer-based methods struggle to learn the interaction between two hands as they directly utilize two hand features as input tokens, which results in distant token problem.
The distant token problem represents that input tokens are in heterogeneous spaces, leading Transformer to fail in capturing correlation between input tokens.
Previous Transformer-based methods suffer from the problem especially when poses of two hands are very different as they project features from a backbone to separate left and right hand-dedicated features.
We present EANet, extract-and-adaptation network, with EABlock, the main component of our network.
Rather than directly utilizing two hand features as input tokens, our EABlock utilizes two complementary types of novel tokens, SimToken and JoinToken, as input tokens.
Our two novel tokens are from a combination of separated two hand features; hence, it is much more robust to the distant token problem.
Using the two type of tokens, our EABlock effectively extracts interaction feature and adapts it to each hand.
The proposed EANet achieves the state-of-the-art performance on 3D interacting hands benchmarks.
The codes are available at \href{https://github.com/jkpark0825/EANet}{https://github.com/jkpark0825/EANet}.
\end{abstract}

\section{Introduction}

\begin{figure}[t]
\begin{center}
\captionsetup{justification=centering}
\captionsetup[sub]{font=scriptsize}
\begin{minipage}{1.\linewidth}
\includegraphics[width=1.\linewidth]
{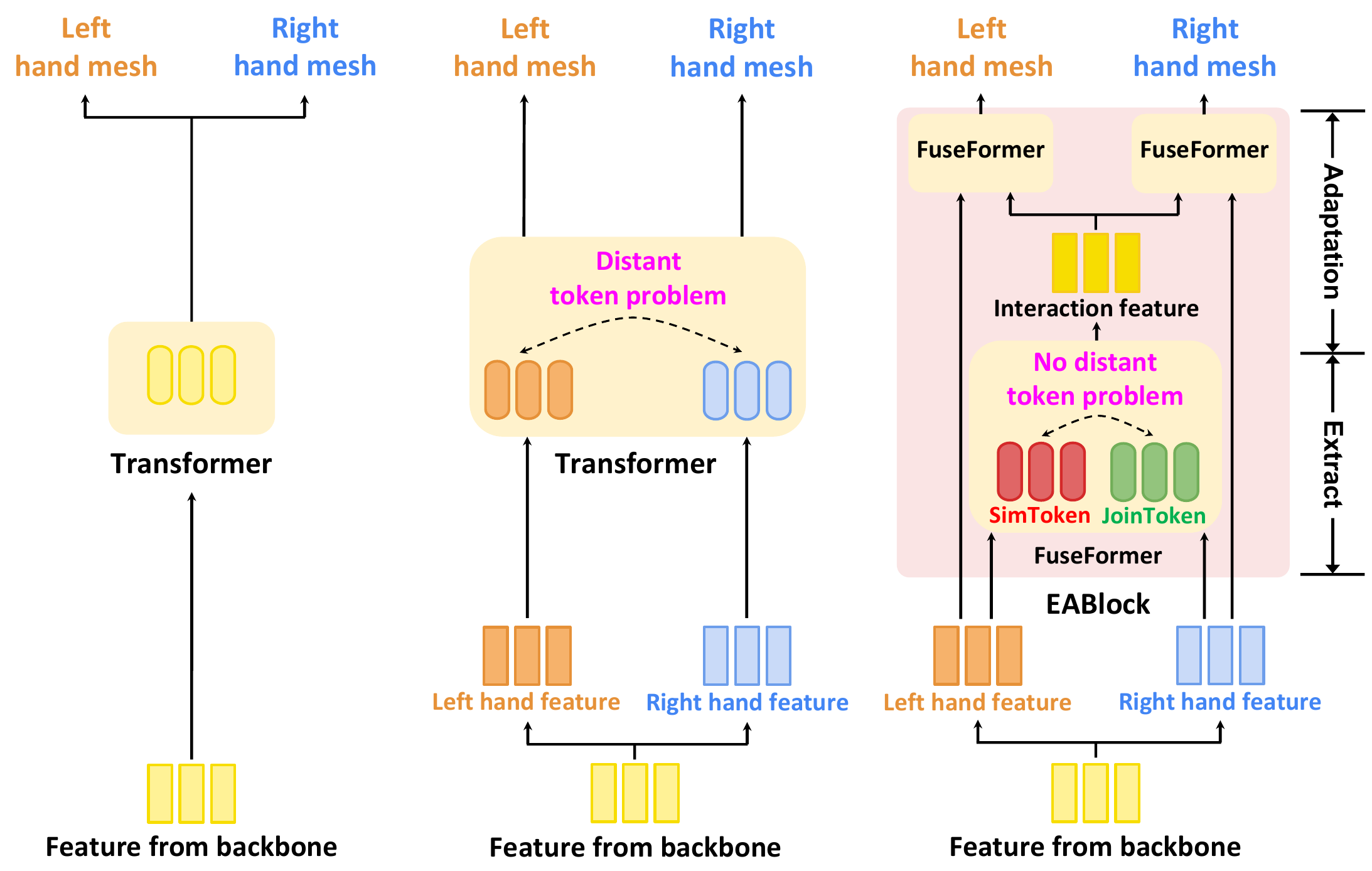}
\begin{minipage}{0.29\linewidth}
\captionsetup[subfigure]{}
\centering
\vspace{-1mm}
\centering
\subcaption{Keypoint Trans~\cite{hampali2022keypoint}\protect\\
\scalebox{1.0}{\textbf{Separate L/R feat.: \xmark}}\protect\\
\scalebox{1.0}{\textbf{No distant token: \cmark}}\label{fig:firstshowing3a}}
\end{minipage}
\begin{minipage}{0.06\linewidth}
\end{minipage}
\begin{minipage}{0.3\linewidth}
\centering
\captionsetup[subfigure]{}
\vspace{-1mm}
\subcaption{IntagHand~\cite{li2022interacting}
\\
\scalebox{1.0}{\,\,\, \textbf{Separate L/R feat.: \cmark}}
\\
\scalebox{1.0}{\,\,\,\, \textbf{No distant token: \xmark}}}
\label{fig:firstshowing3b}
\end{minipage}
\begin{minipage}{0.03\linewidth}
\end{minipage}
\begin{minipage}{0.35\linewidth}
\centering
\captionsetup[subfigure]{}
\vspace{-1mm}
\subcaption{\textbf{EABlock~(Ours)}
\\
\scalebox{1.0}{\, \textbf{Separate L/R feat.: \cmark}}
\\
\scalebox{1.0}{\, \textbf{No distant token: \cmark}}}
\label{fig:firstshowing3c}
\end{minipage}
\end{minipage}
\end{center}
\vspace{-6mm}
\caption{
\textbf{Comparison between previous Transformer blocks and our EABlock.} 
Instead of directly using \textcolor{shared_hand_yellow}{feature from backbone} or \textcolor{left_hand_orange}{left} and \textcolor{right_hand_blue}{right} hand features as input tokens, our FuseFormer in EABlock uses \textcolor{simtoken_red}{SimToken} and \textcolor{jointoken_green}{JoinToken} as input tokens.}
\vspace{-3mm}
\label{fig:firstshowing3}
\end{figure}
\begin{figure}[t]
\begin{center}
\captionsetup[sub]{font=scriptsize}
\begin{minipage}{0.96\linewidth}
\centering
\includegraphics[width=1.0\linewidth]{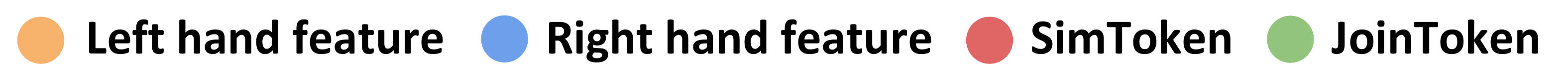
}\\
\includegraphics[width=0.052\linewidth]{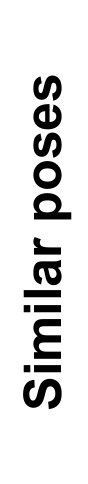}
\includegraphics[width=0.3\linewidth]{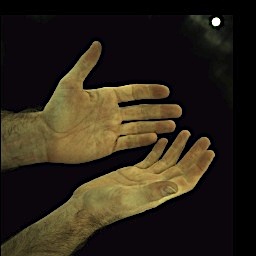}
\includegraphics[width=0.3\linewidth]{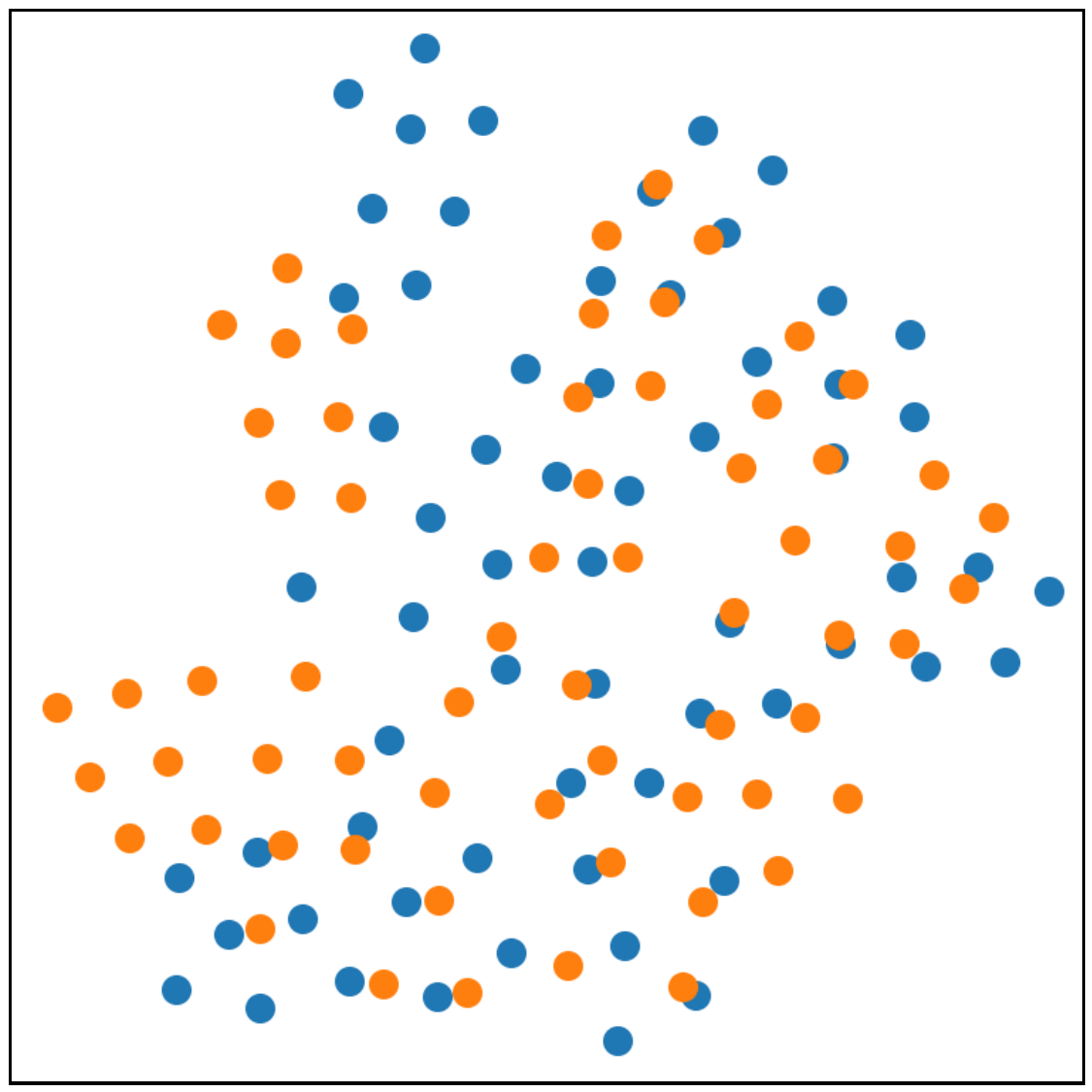}
\includegraphics[width=0.3\linewidth]{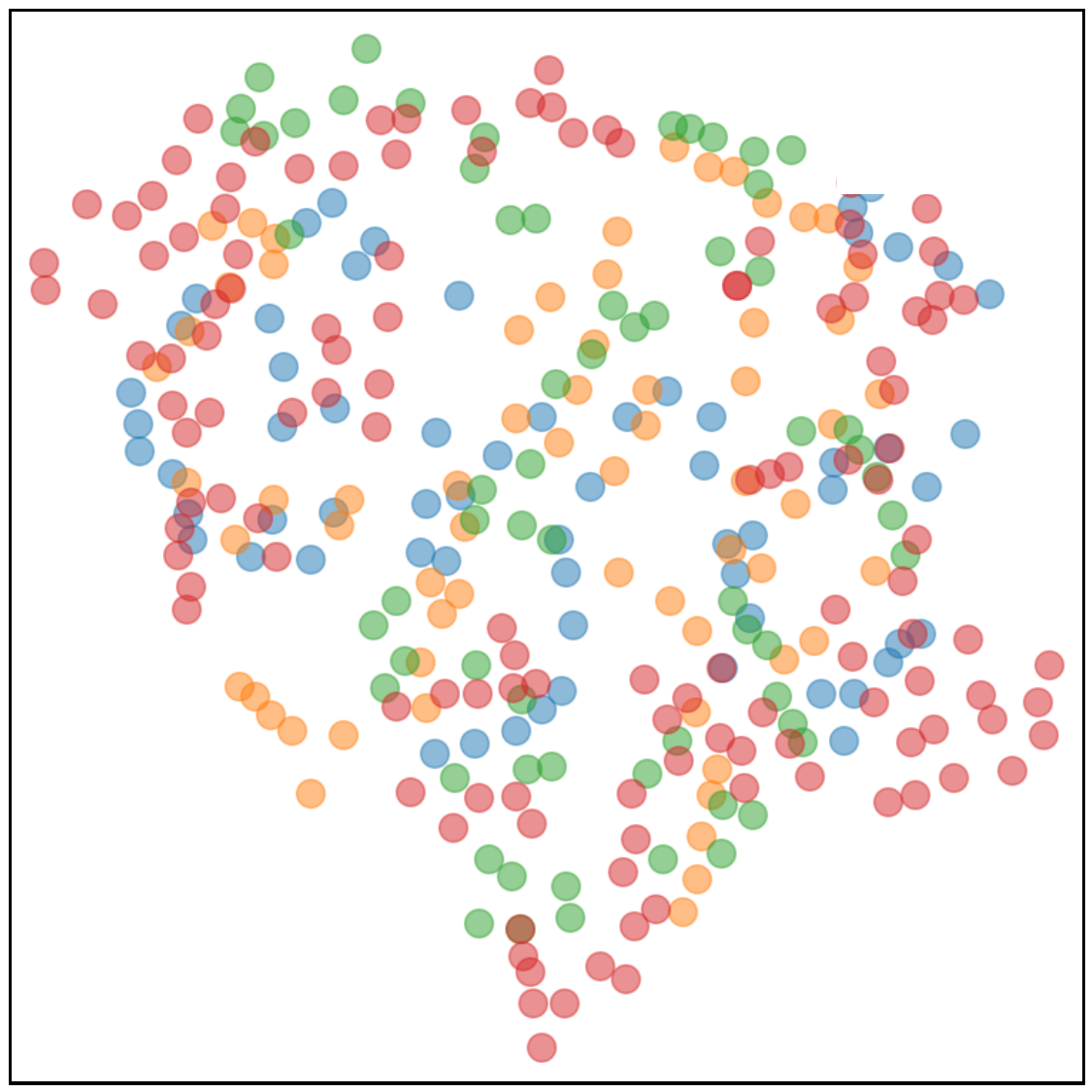}
\\
\includegraphics[width=0.052\linewidth]{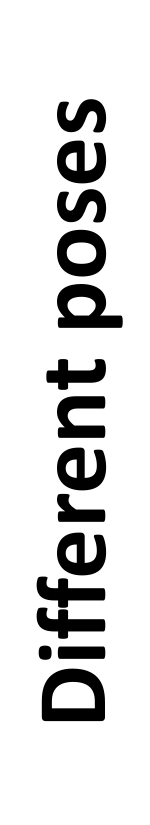}
\includegraphics[width=0.3\linewidth]{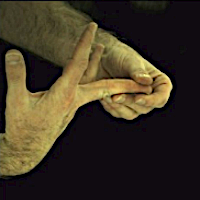}
\includegraphics[width=0.3\linewidth]{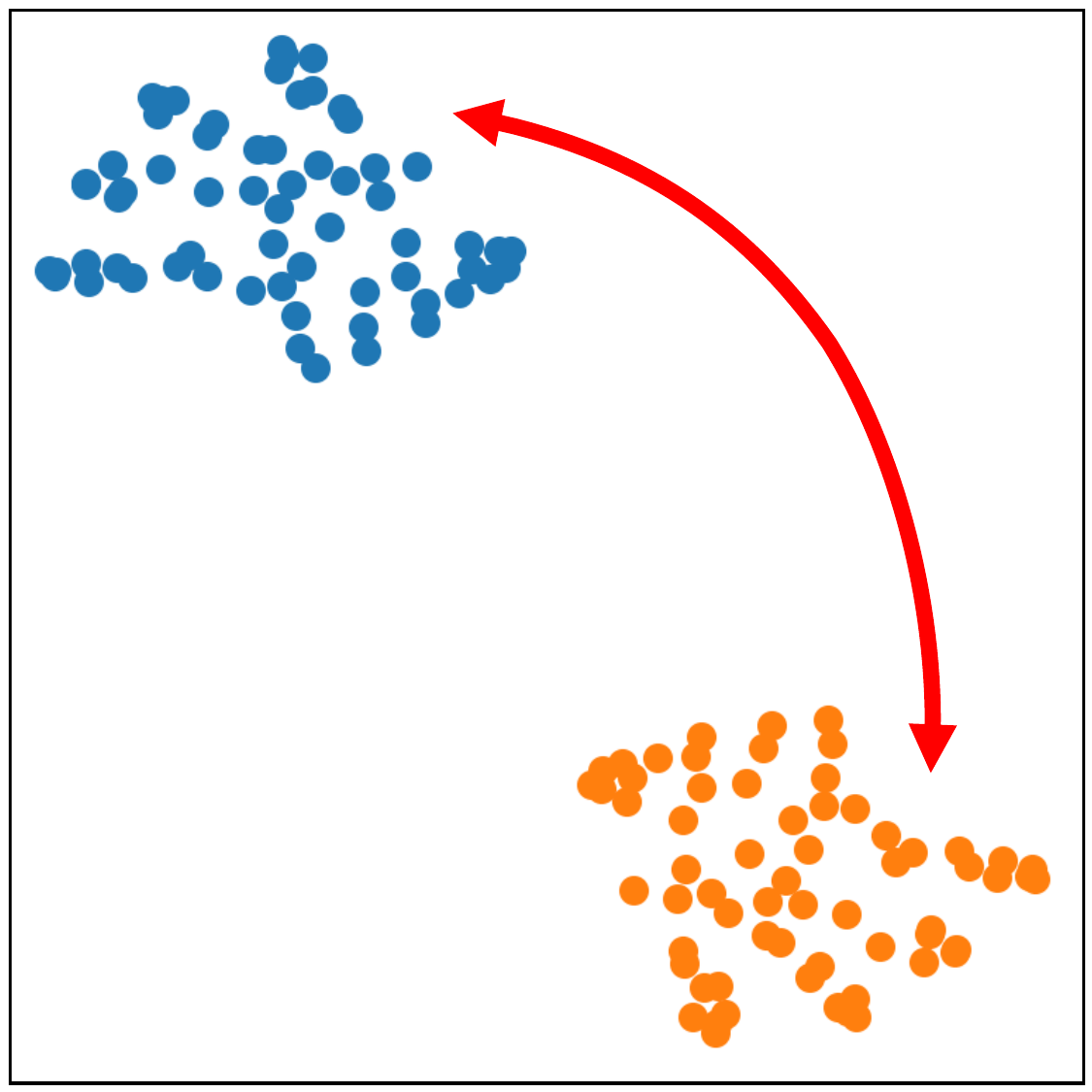}
\includegraphics[width=0.3\linewidth]{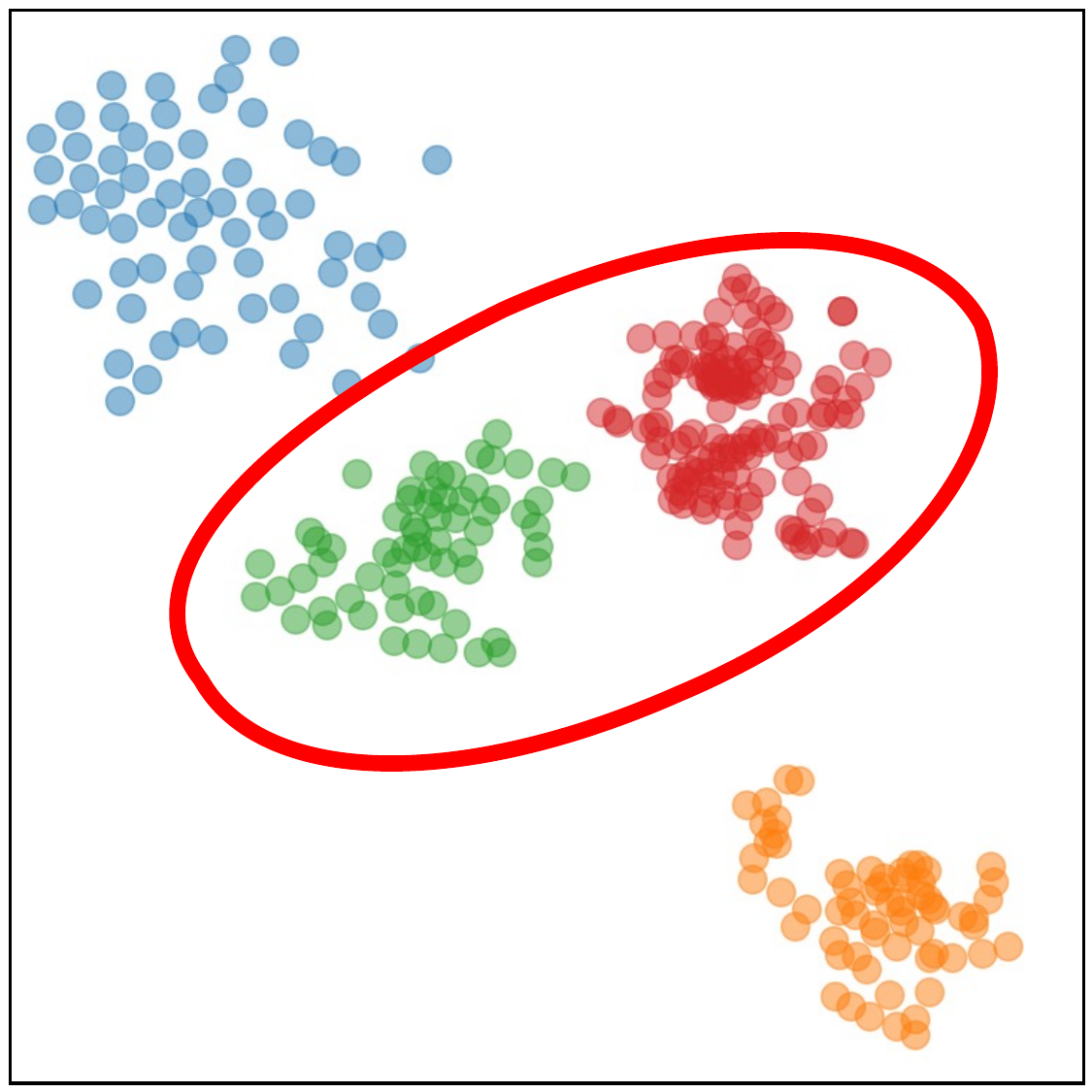}
\\
\vspace{-2mm}
\centering
\begin{minipage}{0.04\linewidth}
\captionsetup[subfigure]{labelformat=empty}
\subcaption{}
\end{minipage}
\addtocounter{subfigure}{-1}
\begin{minipage}{0.29\linewidth}
\captionsetup[subfigure]{}
\subcaption{Input image \label{fig:firstshowinga}}
\end{minipage}
\begin{minipage}{0.3\linewidth}
\captionsetup[subfigure]{}
\subcaption{IntagHand~\cite{li2022interacting} \label{fig:firstshowingb}}
\end{minipage}
\begin{minipage}{0.3\linewidth}
\captionsetup[subfigure]{}
\subcaption{\textbf{EANet~(Ours)} \label{fig:firstshowingc}}
\end{minipage}
\end{minipage}
\centering
\end{center}
\vspace{-6mm}
\caption{\textbf{Comparison of token distribution with t-SNE~\cite{van2008visualizing}}. 
t-SNEs are drawn with input tokens, when given a single image of (a).
(b) Previous methods~\cite{li2022interacting} use \textcolor{left_hand_orange}{left} and \textcolor{right_hand_blue}{right} hand features as input tokens, which suffer from the distant token problem when poses of two hands are different.
(c) We use \textcolor{simtoken_red}{SimToken} and \textcolor{jointoken_green}{JoinToken} as input tokens, suffering much less from the problem.
}
\vspace{-3mm}
\label{fig:firstshowing2}
\end{figure}

Recovering 3D meshes from two interacting hands~\cite{moon2020interhand2, zhang2021interacting, kim2021end, fan2021learning, hampali2022keypoint, li2022interacting} is necessary for immersive experiences in AR/VR and human-computer interaction.
Although the interactions between two hands are highly prevalent in the real world, building a robust 3D mesh recovery framework for two interacting hands is still an open challenge.

The main challenge for recovering the mesh of interacting hands lies in capturing the context of their interaction, which differs from recovering a single hand mesh.
Recent works~\cite{hampali2022keypoint,li2022interacting,di2023lwa}, inspired by the success of the attention mechanism~\cite{vaswani2017attention,dosovitskiy2020image}, use Transformer~\cite{vaswani2017attention} to capture the correlation between the two hands.
There have been two main approaches.
Figure~\ref{fig:firstshowing3a} shows the first approach~\cite{hampali2022keypoint}.
It does not separate features, extracted from a backbone~\cite{he2016deep}, to each hand.
Instead, it directly uses features from the backbone as input tokens of a Transformer.
Figure~\ref{fig:firstshowing3b} shows the second approach~\cite{li2022interacting,di2023lwa}.
It separates features from a backbone to left and right hand-dedicated features.
Then, it uses the separated features as input tokens of a Transformer.

Separating the feature from the backbone can make the network robust to the similarity between two hands.
In addition, left and right hand-dedicated branches could focus only on one type of hand instead of two hands, which could relieve burden of modules~\cite{moon2020interhand2,li2022interacting}.
However, it can cause a \emph{distant token problem}.
Keypoint Transformer~\cite{hampali2022keypoint} in Figure~\ref{fig:firstshowing3a} does not separate features, hence it does not suffer from the distant token problem; however, it cannot enjoy the benefit of the separation.
In contrast, IntagHand~\cite{li2022interacting} in Figure~\ref{fig:firstshowing3b} suffers from the distant token problem as it separates features, while enjoying the benefit of the separation.
The distant token problem arises due to the heterogeneous nature of input tokens projected to two separate spaces, causing the Transformer to struggle in capturing correlations between two hands.
Figure~\ref{fig:firstshowingb} demonstrates the existence of the distant token problem between two hand features.
Interestingly, we observed that the degree of heterogeneity between the two hand features becomes severe when the poses of the two hands are significantly different.
The interaction between the two hands can still be important even when the poses of the hands are very different; however, previous works fail to capture such interaction effectively because of the heterogeneity of the two hand features.
As a result, this leads to sub-optimal performance.
Such distant token problem has also been addressed in the recent multi-modal learning community~\cite{nagrani2021attention, jaegle2021perceiver, liang2021attention, jaegle2021perceiverio}.
While the distant token problem of two hands may not be as severe as in the case of multi-modality, it is still a challenge that needs to be addressed.

To address the distant token problem, we newly introduce two tokens: SimToken and JoinToken, which lie in homogeneous spaces, as shown in Figure~\ref{fig:firstshowingc}, containing two hand information.
Instead of directly passing the separated left and right hand features to Transformers like previous works~\cite{li2022interacting,di2023lwa}, we convert the left hand and right hand features to new tokens and pass the converted ones to Transformers.
The SimToken is obtained through \textbf{sim}ilarity-based operation using a self-attention (SA) Transformer~\cite{vaswani2017attention}.
The JoinToken is obtained by forwarding a concatenation of two input features to a fully connected layer; therefore, JoinToken models \textbf{join}t distribution of two input features without any similarity-based operations.
As the new tokens are from both left and right hand features, they lie in homogeneous spaces.
We design the new tokens to be complementary.
For example, when two hands have very different poses, the distant token problem (the left and right hand features from bottom row of Figure~\ref{fig:firstshowingc}) limits the ability of SA Transformer~\cite{vaswani2017attention} to capture their interaction.
This can result in making SimToken less effective.
On the other hand, JoinToken can still capture useful interaction information in such cases as it does not rely on similarity-based operations (\textit{e.g.}, dot products in Transformer).
Conversely, when two hands have similar poses, SimToken's SA mechanism enables it to capture highly useful interaction information that complements JoinToken.
In this way, the new two tokens allow our system to 1) be robust to similarity between two hands by separating left and right hand features and 2) avoid the distant token problem and compute the correlation between input tokens properly.

We utilize the newly introduced two tokens in EANet, extract-and-adaptation network, where EABlock forms a main component of it.
Figure~\ref{fig:firstshowing3c} briefly shows EABlock, driven by our Transformer-based module, FuseFormer.
FuseFormer fuses two input features by generating the two newly introduced tokens, SimToken and JoinToken, and processing them with a cross-attention (CA) Transformer.

Driven by the FuseFormer, the EABlock consists of extract and adaptation stages.
In the extract stage, our EABlock first \emph{extracts} an interaction feature from two hand features with a FuseFormer.
In the adaptation stage, we \emph{adapt} the extracted interaction feature to each hand with two additional FuseFormers for each hand.
For the left hand adaptation as an example, we pass the extracted interaction feature and left hand feature to FuseFormer for fusing them.
As the interaction feature mostly contains information about both hands, it may not be optimal for separately recovering the 3D hand mesh of each hand.
Therefore, we fuse the interaction feature and left hand feature to achieve two objectives: 1) preserving the interaction information between the two hands and 2) obtaining more left hand-specific information.
We follow a similar process for the right hand adaptation, passing the extracted interaction and right hand features through another FuseFormer.

With extensive experiments and qualitative analysis, we demonstrate the effectiveness of our framework. 
In the challenging InterHand2.6M~\cite{moon2020interhand2} and HIC~\cite{tzionas2016capturing} datasets, we show that our EANet outperforms the previous state-of-the-art methods by a significant margin.
Our contributions are summarized as follows:
\begin{itemize}
\item We propose EANet, extract-and-adaptation network, for 3D interacting hand mesh recovery.
With the help of main block, EABlock, our system effectively captures the interaction between two hands even when poses of two hands are very different.

\item We design FuseFormer, a core component of our EABlock.
FuseFormer fuses input features by generating two complementary types of tokens, SimToken and JoinToken.

\item Extensive experiments and qualitative analysis show that the proposed EANet outperforms previous state-of-the-art methods on 3D hand mesh benchmarks.
\end{itemize}

\section{Related works}

\noindent\textbf{3D interacting hand mesh recovery.}
Several works~\cite{zimmermann2017learning, zhou2020monocular, boukhayma20193d, moon2020interhand2, zhang2021interacting, hampali2022keypoint, li2022interacting} have been proposed to address 3D hand mesh recovery from monocular RGB image.
InterHand2.6M dataset~\cite{moon2020interhand2,moon2022neuralannot} is the first dataset to show a potential for 3D hand mesh recovery in interacting hands by proposing the first large-scale high-resolution real RGB-based 3D hand pose dataset.
Following the release of the dataset~\cite{moon2020interhand2}, Zhang~\etal~\cite{zhang2021interacting} proposed a method that estimates 3D hand meshes based on initially estimated 2.5D heatmap of interacting two hand joints.
The initially estimated 3D hand meshes are then jointly refined in a cascaded manner to take advantage of multi-scale features extracted from pre-trained ResNet-50~\cite{he2016deep} backbone.
As the Transformer~\cite{vaswani2017attention}-based mechanisms have emerged for various tasks in computer vision~\cite{dosovitskiy2020image, he2022masked}, several methods~\cite{li2022interacting, hampali2022keypoint} have been proposed to effectively exploit Transformer~\cite{vaswani2017attention} in 3D interacting hand mesh recovery.
Keypoint Transformer~\cite{hampali2022keypoint} introduced a SA Transformer~\cite{vaswani2017attention}-based module that performs global context-aware feature encoding for keypoints features.
Stepping forward, IntagHand~\cite{li2022interacting} proposed a CA Transformer~\cite{vaswani2017attention}-based module to inherently model two hand correlation between two interacting hands.

Despite promising results, previous Transformer-based methods~\cite{li2022interacting,di2023lwa} suffer from the distant token problem as they directly utilize two hand features as input tokens when poses of two hands are different.
In contrast, our EANet is designed to address such distant token problem by using our two novel tokens as input tokens: SimToken and JoinToken.
The two tokens are prepared and processed in our FuseFormer.
By employing multiple FuseFormers in our main block, EABlock, the proposed EANet effectively \emph{extracts} interaction feature and \emph{adapts} it to each hand.

\noindent\textbf{Tokenization in Transformer.}
Transformer~\cite{vaswani2017attention}-based methods have become the de-facto standards in multiple tasks for both natural language processing~\cite{devlin2018bert, brown2020language} and vision~\cite{dosovitskiy2020image, carion2020end, he2022masked}.
The rise of Transformer~\cite{vaswani2017attention} has largely been accounted for its general purpose inductive bias, allowing flexibility~\cite{jaegle2021perceiver, arnab2021vivit, liu2021fuseformer} to handle input data while making minimal assumptions~\cite{jaegle2021perceiver}.
However, to compensate for such flexibility, many works have demonstrated task-specific add-ons such as positional embedding~\cite{shiv2019novel, jaegle2021perceiverio, likhomanenko2021cape} or tokenization~\cite{dosovitskiy2020image, gorishniy2021revisiting, arnab2021vivit, yuan2021tokens, liu2021fuseformer}.
Especially, tokenization~\cite{dosovitskiy2020image, arnab2021vivit, yuan2021tokens, liu2021fuseformer} has been widely developed in vision as tokens in Transformer do not directly correspond to any single structure in an image.
As demonstrated in vision, proper tokenization is essential for Transformer to make its full potential in corresponding tasks~\cite{zeng2022not,yan2022addressing}.
In our work, we introduce two novel tokens named SimToken and JoinToken.
The two novel tokens allow Transformer to overcome distant token problem of two input features by effectively fusing the input features.
\begin{figure*}[t]
\begin{center}
\includegraphics[width=0.84\linewidth]{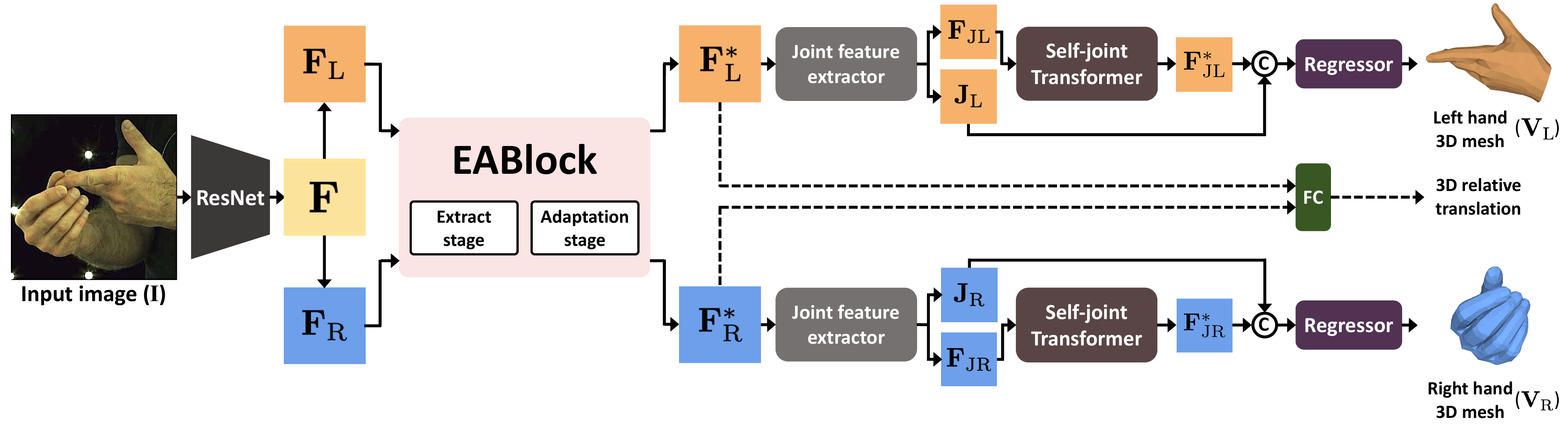}
\end{center}
\vspace{-6mm}
\caption{\textbf{The overall architecture of extract-and-adaptation network (EANet).} 
Our EANet \emph{extracts} an interaction feature and \emph{adapts} it to each hand.
Then, from the adapted features ($\mathbf{F}_\text{L}^*$ or $\mathbf{F}_\text{R}^*$), joint feature extractors extract joint features of each hand ($\mathbf{F}_{\text{JL}}$ or $\mathbf{F}_{\text{JR}}$), guided by corresponding predicted 2.5D joint coordinates ($\mathbf{J}_{\text{L}}$ or $\mathbf{J}_{\text{R}}$).
Finally, regressor produces MANO parameters, which are forwarded to MANO layers for reconstructing 3D hand meshes ($\mathbf{V}_{\text{L}}$ or $\mathbf{V}_{\text{R}}$).
© denotes concatenation.}
\vspace{-3mm}
\label{fig:model}
\end{figure*}

\begin{figure}[t]
\begin{center}
\includegraphics[width=0.8\linewidth]{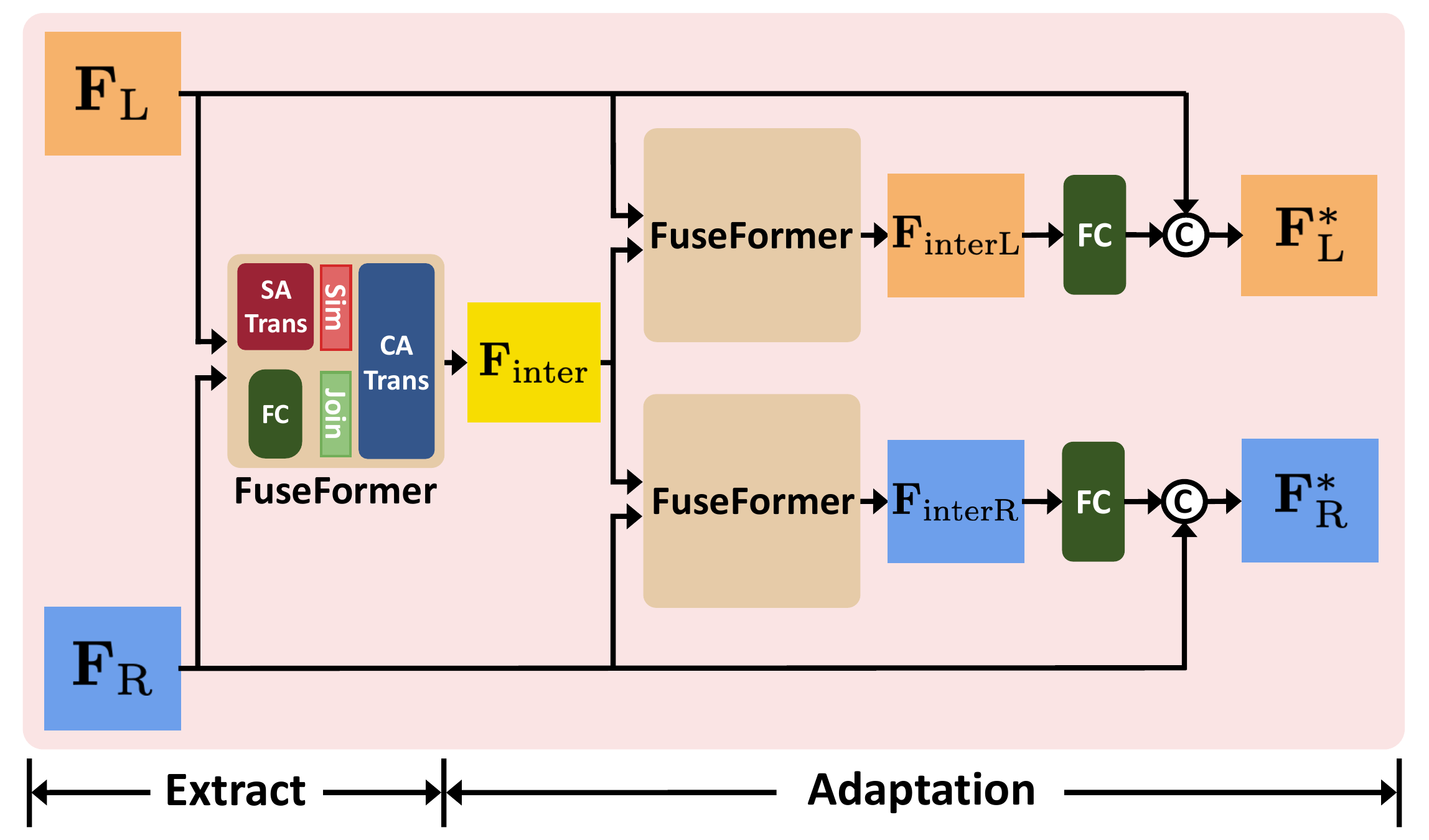}
\end{center}
\vspace{-5mm}
\caption{\textbf{The overall pipeline of EABlock.} A FuseFormer in the extract stage of EABlock first extracts an interaction feature~$\mathbf{F}_{\text{inter}}$ by using both hand features~$\mathbf{F}_{\text{L}}$ and $\mathbf{F}_{\text{R}}$. 
Then, with an interaction feature~$\mathbf{F}_{\text{inter}}$ and one hand feature~($\mathbf{F}_{\text{L}}$ or $\mathbf{F}_{\text{R}}$), each FuseFormer in the adaptation stage of EABlock produces adapted interaction feature~($\mathbf{F}_{\text{interL}}$ or $\mathbf{F}_{\text{interR}}$). 
Afterwards, adapted interaction features are concatenated with corresponding hand features to produce adapted hand feature ($\mathbf{F}_\text{L}^*$ or $\mathbf{F}_\text{R}^*$).
© denotes concatenation.}
\vspace{-3mm}
\label{fig:intt}
\end{figure}

\section{Extract-and-adaptation network~(EANet)}
Figure~\ref{fig:model} provides an overall architecture of our EANet.
The proposed EANet consists of backbone, EABlock, joint feature extractor, self-joint Transformer, and regressor.

\subsection{Backbone}
Given an input hand image~$\mathbf{I} \in \mathbb{R}^{H \times W \times 3}$, a backbone extracts hand features for both left hand~$\mathbf{F}_{\text{L}}$ and right hand~$\mathbf{F}_{\text{R}}$.
$H=256$ and $W=256$ represent height and width of the input image, respectively.
We first feed a hand image~$\mathbf{I}$ to ResNet-50~\cite{he2016deep}, pre-trained on ImageNet~\cite{deng2009imagenet}, to extract an image feature~$\mathbf{F} \in \mathbb{R}^{h\times w\times C}$, where $h=H/32$ and $w=W/32$ denote height and width of $\mathbf{F}$ respectively, and $C=2048$ denotes the channel dimension of $\mathbf{F}$.
Then, the image feature $\mathbf{F}$ is passed to two separate $1\times1$ convolutional layers to obtain left hand feature~$\mathbf{F}_{\text{L}}$ and right hand feature~$\mathbf{F}_{\text{R}}$.
The two features have the same dimension of $\mathbb{R}^{h\times w\times c}$ where $c=C/4$.

\subsection{Extract-and-adaptation block~(EABlock)~\label{sec:eablock}}
Figure~\ref{fig:intt} shows detailed pipeline of our EABlock.
EABlock consists of two stages~(\ie, extract and adaptation), and FuseFormers are used in each stage.
FuseFormer is our basic block, which fuses two types of input features with SimToken and JoinToken.
FuseFormer consists of a SA Transformer, a fully connected layer, and a CA Transformer.

\begin{figure}[t]
\begin{center}
\hspace{2mm}
\includegraphics[width=0.85\linewidth]{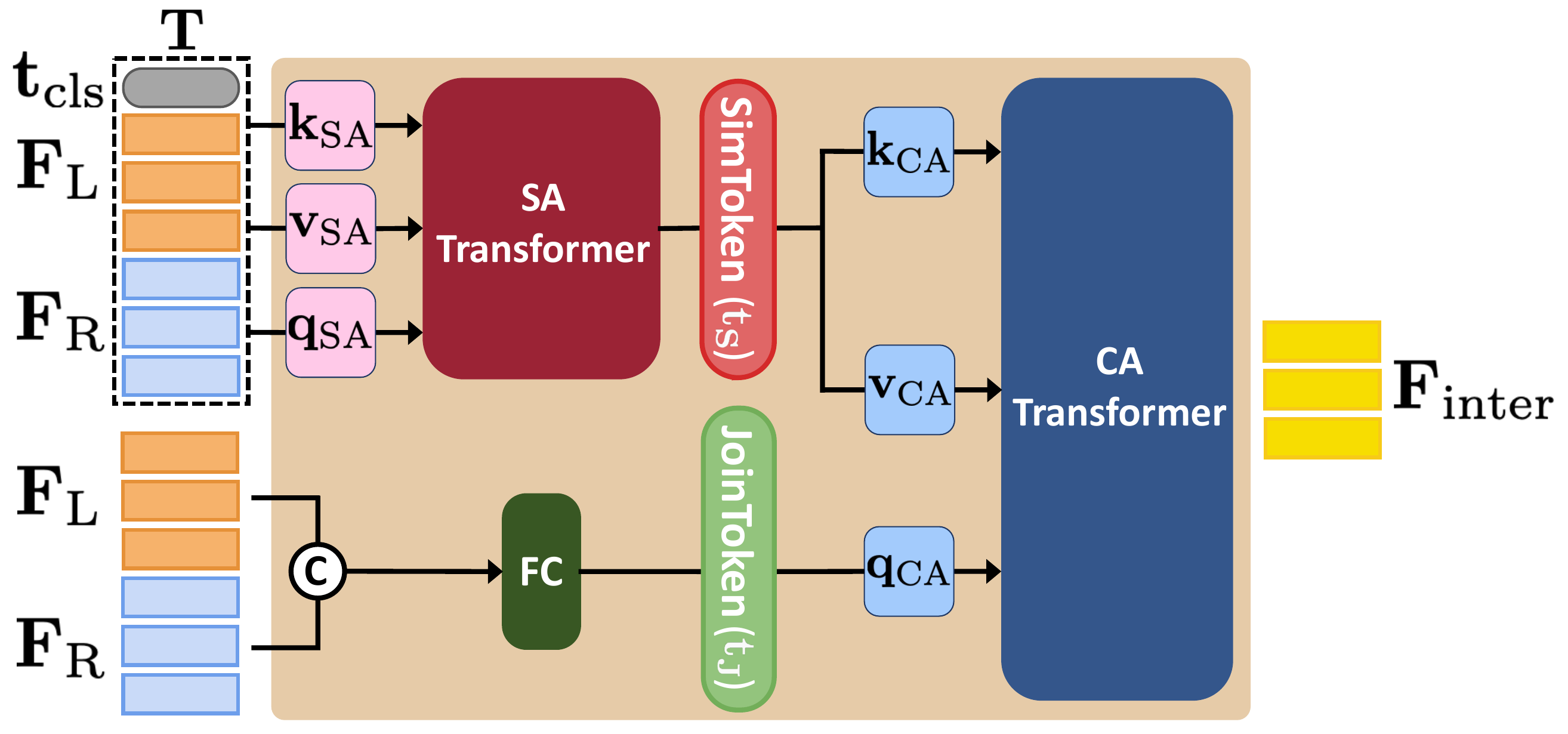}
\end{center}
\vspace{-4mm}
\caption{\textbf{The overall pipeline of FuseFormer in extract stage.}
FuseFormer first prepares \textcolor{simtoken_red}{SimToken~$\mathbf{t}_{\text{S}}$} and \textcolor{jointoken_green}{JoinToken~$\mathbf{t}_{\text{J}}$} with SA Transformer and a fully connected layer, respectively.
Then, a CA Transformer processes the two tokens.
© denotes concatenation.
}
\vspace{-3mm}
\label{fig:inttb}
\end{figure}

\noindent \textbf{Extract: Interaction feature extract.}
Figure~\ref{fig:inttb} shows the overall pipeline of FuseFormer in the extract stage.
The FuseFormer in the extract stage \emph{extracts} an interaction feature~$\mathbf{F}_{\text{inter}}$ by fusing two hand features ($\mathbf{F}_{\text{L}}$ and $\mathbf{F}_{\text{R}}$).
Different from the previous Transformer-based methods~\cite{hampali2022keypoint,li2022interacting} that utilize $\mathbf{F}_{\text{L}}$ and $\mathbf{F}_{\text{R}}$ as input tokens, we use SimToken~$\mathbf{t}_{\text{S}}$ and JoinToken~$\mathbf{t}_{\text{J}}$ as input tokens to address the distant token problem.

SimToken~$\mathbf{t}_{\text{S}}$ is obtained by passing the two hand features ($\mathbf{F}_{\text{L}}$ and $\mathbf{F}_{\text{R}}$) to a SA Transformer~\cite{vaswani2017attention}.
To this end, we first reshape $\mathbf{F}_{\text{L}}$ and $\mathbf{F}_{\text{R}}$ from $\mathbb{R}^{h\times w\times c}$ to  $\mathbb{R}^{hw \times c}$.
Then, we concatenate the reshaped $\mathbf{F}_{\text{L}}$ and $\mathbf{F}_{\text{R}}$ along with a class token $\mathbf{t}_{\text{cls}} \in \mathbb{R}^{1\times c}$~\cite{dosovitskiy2020image}.
The class token is implemented as a learnable one-dimensional embedding vector to learn a general two hand information~\cite{dosovitskiy2020image}.
We denote the concatenated token by $\mathbf{t} \in \mathbb{R}^{l \times c}$, where $l = hw+hw+1$.
Then, we extract query~$\mathbf{q_\text{SA}}$, key~$\mathbf{k_\text{SA}}$, and value~$\mathbf{v_\text{SA}}$ from the concatenated token~$\mathbf{t}$ with separate linear layers.
The dimensions of $\mathbf{q_\text{SA}}$, $\mathbf{k_\text{SA}}$, and $\mathbf{v_\text{SA}}$ are all identical to that of $\mathbf{t}$.
Analogous to the previous Transformers~\cite{vaswani2017attention,dosovitskiy2020image}, the SimToken~$\mathbf{t}_{\text{S}} \in \mathbb{R}^{l \times c}$ is produced as follows:
\begin{equation}
\text{Attn}(\mathbf{q_\text{SA}},\mathbf{k_\text{SA}},\mathbf{v_\text{SA}}) = \text{softmax}(\frac{{\mathbf{q_\text{SA}}}{\mathbf{k_\text{SA}}}^{T}}{\sqrt{d_\mathbf{k_\text{SA}}}})\mathbf{v_\text{SA}},
\label{eq:1}
\end{equation}
\begin{equation}
\mathbf{r} = \mathbf{t} +\text{Attn}(\mathbf{q_\text{SA}},\mathbf{k_\text{SA}},\mathbf{v_\text{SA}}),
\label{eq:2}
\end{equation}
\begin{equation}
\mathbf{t}_{\text{S}} = \mathbf{r} + \text{MLP}(\mathbf{r}),
\label{eq:3}
\end{equation}
where $d_\mathbf{k_\text{SA}}$ denotes a channel dimension of $\mathbf{k_\text{SA}}$.
Consequently, $\mathbf{t}_{\text{S}}$ is built upon a standard SA Transformer from two hand features.

JoinToken~$\mathbf{t}_{\text{J}} \in \mathbb{R}^{hw\times c}$ is obtained by passing two hand features ($\mathbf{F}_{\text{L}}$ and $\mathbf{F}_{\text{R}}$) to a fully connected layer.
Before passing the two hand features, we concatenate and reshape two hands features ($\mathbf{F}_{\text{L}}$ and $\mathbf{F}_{\text{R}}$) from $\mathbb{R}^{h \times w \times 2c}$ to $\mathbb{R}^{hw\times 2c}$.
Formally,
\begin{equation}
\mathbf{t}_{\text{J}} = \text{FC}(\psi(\mathbf{F}_\text{L}, \mathbf{F}_\text{R})),
\end{equation}
where $\psi$ represents a composition of concatenation and reshape functions.
$\text{FC}$ denotes a fully connected layer.

After preparing two tokens ($\mathbf{t}_{\text{S}}$ and $\mathbf{t}_{\text{J}}$), a CA~\cite{chen2021crossvit} Transformer extracts an interaction feature~$\mathbf{F}_{\text{inter}} \in \mathbb{R}^{hw \times c}$ from the two tokens.
For CA, we extract query from JoinToken~$\mathbf{t}_{\text{J}}$ and key-value pair from SimToken~$\mathbf{t}_{\text{S}}$ with separate linear layers.
We denote query, key and value of the CA Transformer by $\mathbf{q}_\text{CA} \in \mathbb{R}^{hw \times c}$, $\mathbf{k}_\text{CA} \in \mathbb{R}^{l \times c}$, and $\mathbf{v}_\text{CA} \in \mathbb{R}^{l \times c}$, respectively.

The reason for using JoinToken as a query is to address \emph{mismatch between the query-key correlation and value}.
Let us take an example where the inputs of the CA Transformer are obtained in an opposite way: query from SimToken and key-value from JoinToken.
When poses of two hands are very different, SimToken fails to capture useful interaction between two hands; hence, first half and second half of SimToken mainly contain the left and right hand information, respectively, not both hand information.
On the other hand, JoinToken contains information of both hands.
Let us take a single element in query, which mostly contains left hand information, as an example.
From this element's point of view, the query-key correlation is always formulated by how much all elements in JoinToken are similar to the left hand information.
However, as value is from JoinToken, the opposite hand  information can be retrieved based on the left hand-driven similarity, which is undesirable.
In the end, using SimToken as a query suffers from the mismatch between query-key correlation (driven by the left hand) and value (right hand information).
In contrast, ours extracts query from JoinToken and key-value from SimToken.
As elements in query contain both hand information, the query-key correlation is not fixed to a single type of hand (\textit{e.g.}, left hand); hence, ours does not suffer from the mismatch.

In the end, the CA Transformer outputs an interaction feature~$\mathbf{F}_{\text{inter}}$ following Equation~\ref{eq:1}, \ref{eq:2}, and \ref{eq:3} with $\mathbf{q}_\text{CA}$, $\mathbf{k}_\text{CA}$, and $\mathbf{v}_\text{CA}$.
Consequently, FuseFormer in the extract stage \emph{extracts} interaction feature~$\mathbf{F}_{\text{inter}}$ by fusing two hand features ($\mathbf{F}_{\text{L}}$ and $\mathbf{F}_{\text{R}}$).
Formally, 
\begin{equation}
\mathbf{F}_{\text{inter}} = \text{FuseFormer}(\mathbf{F}_{\text{L}}, \mathbf{F}_{\text{R}}; \mathbf{W}_\text{extract}),
\label{eq:5}
\end{equation}
where $\mathbf{W}_\text{extract}$ denotes learnable weights of the FuseFormer in the extract stage.

\noindent\textbf{Adaptation: Interaction feature adaptation.}
In the adaptation stage, EABlock adapts the extracted interaction feature~$\mathbf{F}_{\text{inter}}$ to each hand with two additional FuseFormers.
While the interaction feature, obtained from the extract stage, is useful for understanding how two hands interact with each other, directly utilizing it for 3D hand mesh recovery of each hand might not be optimal.
Therefore, we fuse the interaction feature and feature of each hand to achieve two goals: 1) preserving interaction information between two hands and 2) obtaining left and right hand-specific information.
Taking the left hand adaptation as an example, we pass the left hand feature $\mathbf{F}_{\text{L}}$ and the interaction feature $\mathbf{F}_{\text{inter}}$ to a FuseFormer, which fuses them and outputs interaction feature adapted to the left hand~$\mathbf{F}_{\text{interL}} \in \mathbb{R}^{hw \times c}$.
The adaptation for the right hand is performed in the same manner by passing the right hand feature $\mathbf{F}_{\text{R}}$ and interaction feature $\mathbf{F}_{\text{inter}}$ to another FuseFormer, which fuses them and outputs interaction feature adapted to the right hand~$\mathbf{F}_{\text{interR}} \in \mathbb{R}^{hw \times c}$.
Formally,
\begin{equation}
\mathbf{F}_{\text{interL}} = \text{FuseFormer}(\mathbf{F}_{\text{L}}, \mathbf{F}_{\text{inter}}; \mathbf{W}_\text{adaptL}),
\label{eq:6}
\end{equation}
\begin{equation}
\mathbf{F}_{\text{interR}} = \text{FuseFormer}(\mathbf{F}_{\text{R}}, \mathbf{F}_{\text{inter}}; \mathbf{W}_\text{adaptR}),
\label{eq:7}
\end{equation}
where $\mathbf{W}_\text{adaptL}$ and $\mathbf{W}_\text{adaptR}$ denote learnable weights in FuseFormers for the left and right hand adaptations, respectively.

To reduce computational costs, for each hand, we apply a fully connected layer, which reduces the channel dimension from $c$ to $c/4$, to the adapted interaction features ($\mathbf{F}_{\text{interL}}$ or $\mathbf{F}_{\text{interR}}$).
Then, we reshape the output of the fully connected layer to $\mathbb{R}^{h \times w \times c/4}$.
Finally, for each hand, we concatenate the adapted interaction feature~($\mathbf{F}_{\text{interL}}$ or $\mathbf{F}_{\text{interR}}$) and its corresponding hand feature~($\mathbf{F}_{\text{L}}$ or $\mathbf{F}_{\text{R}}$) along the channel dimension, which becomes final adapted feature of each hand, denoted by $\mathbf{F}^{*}_{\text{L}}$ and $\mathbf{F}^{*}_{\text{R}}$.

\subsection{Joint feature extractor}
For each hand, a joint feature extractor extracts 2.5D joint coordinates~($\mathbf{J}_{\text{L}}$ or $\mathbf{J}_{\text{R}}$) and joint features~($\mathbf{F}_{\text{JL}}$ or $\mathbf{F}_{\text{JR}}$) from the adapted hand features ($\mathbf{F}^{*}_{\text{L}}$ or $\mathbf{F}^{*}_{\text{R}}$). 
Motivated by Moon~\etal~\cite{moon2022accurate}, our joint feature extractor effectively extracts joint features with the guidance of estimated 2.5D joint coordinates.

\noindent \textbf{2.5D joint coordinate estimation.}
$x$- and $y$-axis of 2.5D joint coordinates are in the 2D pixel space, while $z$-axis of them are in the root joint (\ie, wrist)-relative depth space.
To estimate 2.5D joint coordinates of each hand, we apply a $1\times 1$ convolution layer to the adapted hand features ($\mathbf{F}^{*}_{\text{L}}$ or $\mathbf{F}^{*}_{\text{R}}$) to change their channel dimension to $dJ$.
Then, we reshape the output to a 2.5D heatmap whose dimension is $\mathbb{R}^{h \times w \times d \times J}$.
$d=8$ denotes discretized depth size of the 2.5D heatmap, and $J$ denotes the number of joints.
Subsequently, 2.5D joint coordinates for left or right hand~($\mathbf{J}_{\text{L}}$ or $\mathbf{J}_{\text{R}}$) are obtained by applying the soft-argmax operation~\cite{sun2018integral} to the corresponding 2.5D heatmap.

\noindent \textbf{Joint feature extract.}
The left hand joint features~$\mathbf{F}_{\text{JL}}$ are obtained through a grid sampling~\cite{jaderberg2015spatial} on feature map~$\mathbf{F}^{*}_{\text{L}}$ with $(x,y)$ position from the estimated 2.5D joint coordinates~$\mathbf{J}_{\text{L}}$.
Likewise, the right hand joint features~$\mathbf{F}_{\text{JR}}$ are obtained in a same manner with right hand feature~$\mathbf{F}^{*}_{\text{R}}$ and the corresponding 2.5D joint coordinates~$\mathbf{J}_{\text{R}}$.
The joint features contain global contextual information of each hand joint, essential for 3D hand mesh recovery.

\subsection{Self-joint Transformer (SJT)}
Self-joint Transformer~(SJT) is a standard SA Transformer.
For each hand, it enhances joint features ($\mathbf{F}_{\text{JL}}$ or $\mathbf{F}_{\text{JR}}$) through the SA~\cite{vaswani2017attention}, which outputs ($\mathbf{F}^{*}_{\text{JL}}$ or $\mathbf{F}^{*}_{\text{JR}}$).
Despite its simplicity, we observed that SJT is greatly helpful to enhance the joint features as it can implicitly consider kinematic structure of hand joints through the SA.
Note that the goal of SJT is clearly different from that of FuseFormer: FuseFormer aims to fuse two types of input features, while the SJT aims to enhance a single type of input feature.

\subsection{Final outputs}
For each hand, the regressor produces 48-dimensional pose parameters ($\theta_{\text{L}}$ or $\theta_{\text{R}}$) and 10-dimensional shape parameters ($\beta_{\text{L}}$ or $\beta_{\text{R}}$) of MANO~\cite{romero2022embodied} from the enhanced hand joint features~($\mathbf{F}^{*}_{\text{JL}}$ or $\mathbf{F}^{*}_{\text{JR}}$). 
The pose parameters are obtained by first concatenating the enhanced joint features~($\mathbf{F}^{*}_{\text{JL}}$ or $\mathbf{F}^{*}_{\text{JR}}$) with 2.5D joint coordinates~($\mathbf{J}_{\text{L}}$ or $\mathbf{J}_{\text{R}}$).
Then, we flattened the concatenated features each into a one-dimensional vector.
The final two vectors are passed to two fully connected layers to predict MANO pose parameters.
To obtain the shape parameters, we forward the adapted hand features~($\mathbf{F}^{*}_{\text{L}}$ or $\mathbf{F}^{*}_{\text{R}}$) to a fully connected layer after global average pooling~\cite{lin2013network}.
In the end, the final 3D hand meshes, denoted by $\mathbf{V}_{\text{L}}$ and $\mathbf{V}_{\text{R}}$, are obtained by forwarding the MANO parameters to MANO layers.
In addition, the 3D relative translation between two hands is obtained from the adapted hand features ($\mathbf{F}^{*}_{\text{L}}$ and $\mathbf{F}^{*}_{\text{R}}$).
To this end, we concatenate them and perform global average pooling.
Finally, a fully connected layer is used to output the 3D relative translation between two hands.
To train our model, we minimize loss functions, defined as a weighted sum of L1 distances between estimated and ground truth $\theta$, $\beta$, $\mathbf{J}$, $\mathbf{V}$, and 3D relative translation.

\section{Experiments}
\subsection{Implementation details}
All implementations are done with PyTorch~\cite{paszke2017automatic} under Adam optimizer~\cite{kingma2014adam} and in batch size of 32 per GPU~(trained with four RTX 2080 Ti GPUs).
For training our model on InterHand2.6M~\cite{moon2020interhand2} dataset, we trained our model for 30 epochs, with learning rate annealing at 10th and 15th epochs from the initial learning rate~$1\times 10^{-4}$.

\begin{table}[t]
\def\arraystretch{1.2}
    \centering
    \resizebox{0.9\linewidth}{!}{
    \begin{tabular}{c|c|ccc|ccc|c}
        \Xhline{2\arrayrulewidth}
        \multirow{2}{*}{Separate L/R feature} & \multirow{2}{*}{EABlock} & \multicolumn{3}{c|}{MPJPE} & \multicolumn{3}{c|}{MPVPE} & \multirow{2}{*}{MRRPE}\\[-0.2em] 
        & & Single & Two & All & Single & Two & All & \\
        \hline
        \xmark & \xmark & 14.90 & 15.73 & 15.32 & 11.84 & 14.05 & 12.80 & 22.76\\
        \cmark & \xmark & 12.74 & 14.33 & 13.53 & 12.53 & 11.95 & 12.28 & 23.79\\
        \cmark & \cmark & \textbf{9.62} & \textbf{11.54} & \textbf{10.58} & \textbf{7.86} & \textbf{10.78} & 
        \textbf{9.13} & \textbf{18.82}\\
        \Xhline{2\arrayrulewidth}
    \end{tabular}}
    \vspace{-2mm}
    \caption{\textbf{Comparison of models with various architectures in Figure~\ref{fig:firstshowing3}.}}
    \vspace{-3mm}
    \label{tab:projdesign}
\end{table}

\subsection{Datasets and evaluation metrics}
\noindent\textbf{InterHand2.6M dataset.}
We trained and evaluated EANet on InterHand2.6M dataset.
For the comparison with the previous approaches, we used both single and interacting hand~(SH+IH) images of the H+M split.
For ablation studies, we used SH+IH images of the H split.

\noindent\textbf{HIC dataset.}
To show the generalizability of our proposed EANet, we further showed results on HIC~\cite{tzionas2016capturing} dataset.
Different from InterHand2.6M~\cite{moon2020interhand2} dataset, which has homogeneous background and lighting sources, HIC dataset includes natural lighting with more diverse backgrounds.
Note that the HIC dataset was only used for evaluation.

\noindent\textbf{Evaluation metrics.}
Following the prior arts~\cite{moon2020interhand2,hampali2022keypoint}, we reported mean per joint position error (MPJPE), mean per vertex position error (MPVPE), and mean relative-root position error (MRRPE).
All metrics are in a millimeter scale.

\begin{table}[t]
\def\arraystretch{1.2}
    \centering
    \resizebox{0.9\linewidth}{!}{
    \begin{tabular}{c|ccc|ccc|c}
        \Xhline{2\arrayrulewidth}
         \multirow{2}{*}{Block} & \multicolumn{3}{c|}{MPJPE} & \multicolumn{3}{c|}{MPVPE} & \multirow{2}{*}{MRRPE}\\[-0.2em]
        & Single & Two & All & Single & Two & All & \\
        \hline
        CA Transformer & 12.74 & 14.33 & 13.53 & 12.53 & 11.95 & 12.28 & 23.79\\
        SA Transformer & 11.44 & 13.49 & 12.47 & 9.51 & 12.37 & 10.75 & 25.35\\
        Ours (w/o adaptation)  & 10.66 & 11.81 & 11.25 & 8.37 & 11.24 & 9.61 & 21.00\\
        \textbf{Ours (full)}  & \textbf{9.62} & \textbf{11.54} & \textbf{10.58} & \textbf{7.86} & \textbf{10.78} & \textbf{9.13} & \textbf{18.82}\\
        \Xhline{2\arrayrulewidth}
    \end{tabular}}
    \vspace{-2mm}
    \caption{\textbf{Comparison of models with various Transformer blocks for EABlock.}}
    \vspace{-3mm}
    \label{tab:rdtdesign}
\end{table}

\begin{figure}[!t]
\begin{center}
\captionsetup{justification=centering}
\captionsetup[sub]{font=scriptsize}
\begin{minipage}{1.\linewidth}
\centering
\includegraphics[width=0.18\linewidth]{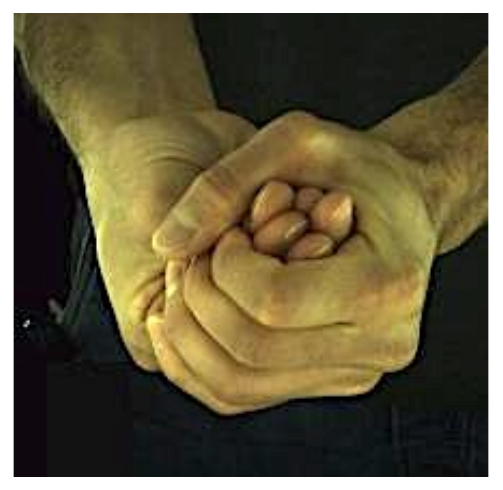}
\includegraphics[width=0.18\linewidth]{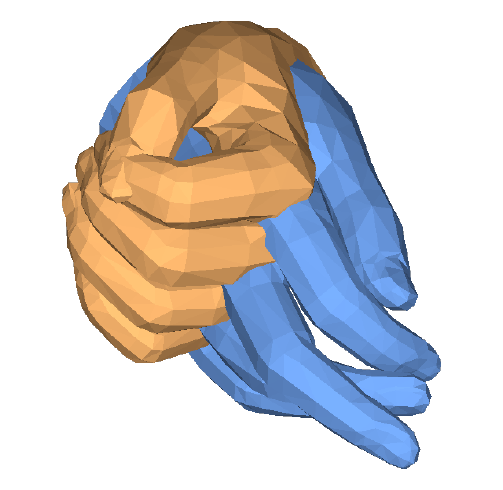}
\includegraphics[width=0.18\linewidth]{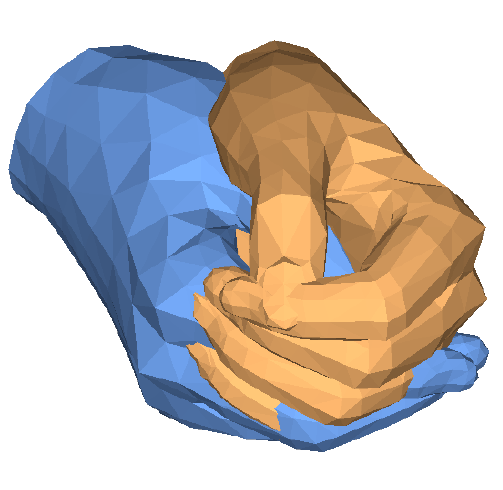}
\includegraphics[width=0.18\linewidth]{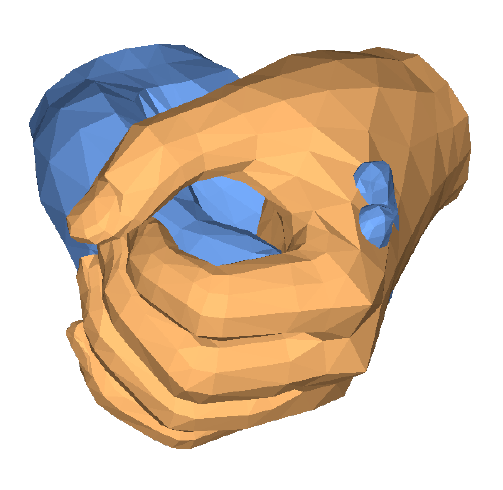}
\includegraphics[width=0.18\linewidth]{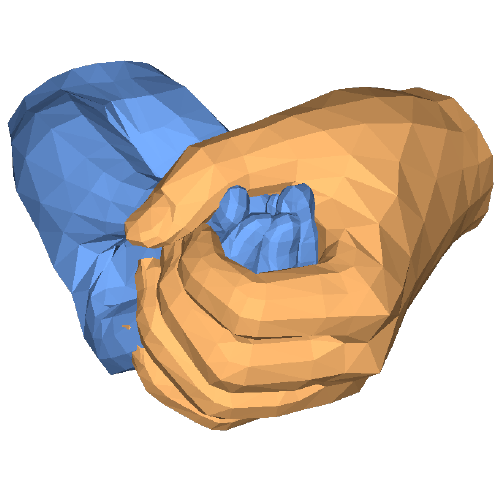}
\\
\vspace{-3mm}
\centering
\begin{minipage}{0.18\linewidth}
\captionsetup[subfigure]{}
\subcaption{Input\\ image \label{fig:qual_abl_img}}
\end{minipage}
\centering
\begin{minipage}{0.18\linewidth}
\captionsetup[subfigure]{}
\subcaption{CA\\ Transformer \label{fig:qual_abl_ca}}
\end{minipage}
\centering
\begin{minipage}{0.18\linewidth}
\captionsetup[subfigure]{}
\subcaption{SA\\ Transformer \label{fig:qual_abl_sa}}
\end{minipage}
\centering
\begin{minipage}{0.18\linewidth}
\captionsetup[subfigure]{}
\subcaption{Ours\\~(w/o adapt.) \label{fig:qual_abl_ours_wo}}
\end{minipage}
\centering
\begin{minipage}{0.18\linewidth}
\captionsetup[subfigure]{}
\subcaption{\textbf{Ours\\~(full)} \label{fig:qual_abl_ours_full}}
\end{minipage}
\end{minipage}
\centering
\end{center}
\vspace{-6mm}
\caption{\textbf{Visual comparison of models with various Transformer blocks for EABlock.}}
\vspace{-3mm}
\label{fig:qual_abl}
\end{figure}

\subsection{Ablation studies}
\begin{figure*}[t]
\captionsetup[sub]{font=scriptsize}
\begin{center}
\includegraphics[width=0.76\linewidth]{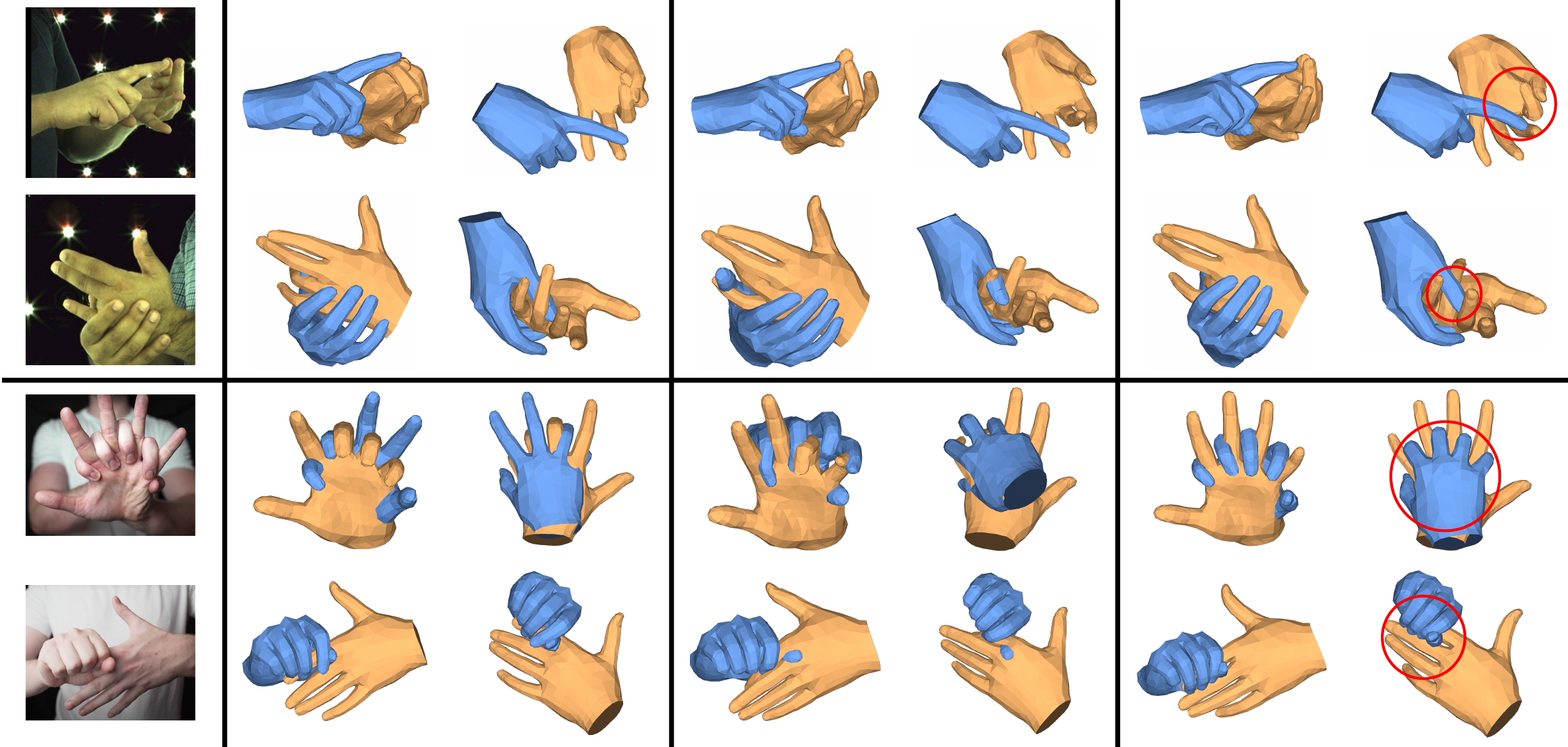}
\\
\vspace{-1mm}
\captionsetup[subfigure]{labelformat=empty}
\begin{minipage}{1.0\linewidth}
\subcaption{\textcolor{white}{AAAA}\textbf{Input image}\textcolor{white}{AAAAAAAA}\textbf{IntagHand}~\cite{li2022interacting}\textcolor{white}{AAAAAAAAA}\textbf{Keypoint Transformer}~\cite{hampali2022keypoint}\textcolor{white}{AAAAAAAAA}\textbf{EANet~(Ours)}\textcolor{white}{AAAAAAAAA}}
\end{minipage}
\end{center}
\vspace{-6mm}
\caption{\textbf{Visual comparison with state-of-the-art methods on InterHand2.6M~\cite{moon2020interhand2}~(top) and in-the-wild~(bottom).}
The red circles highlight regions where our EANet is correct, while others are wrong.
We crawl in-the-wild images from web.
}
\vspace{-3mm}
\label{fig:qualitative}
\end{figure*}

\noindent \textbf{Overall design of EANet.}
Table~\ref{tab:projdesign} justifies our approach, which 1) separates left and right hand features and 2) uses EABlock to address the distant token problem, as depicted in Figure~\ref{fig:firstshowing3c}.
For the first row, we did not separate the feature from the backbone to left and right hand features.
Instead, the feature from the backbone was directly passed to SA Transformer following Keypoint Transformer~\cite{hampali2022keypoint}.
Then, a single joint feature extractor and self-joint Transformer were used.
Two regressors for each hand output final 3D hand meshes.
For the second row, we replaced EABlock with CA Transformer following IntagHand~\cite{li2022interacting} while keeping the separation.
Without the separation (Figure~\ref{fig:firstshowing3a} and the first row of the table), model's performance drops significantly as the separation is helpful to make the network robust to the similarity between left and right hands.
Also, left and right hand-dedicated branches could focus only on one type of hands instead of two hands, which could relieve the burden of modules~\cite{moon2020interhand2,li2022interacting}. 
Compared to the second row of the table (Figure~\ref{fig:firstshowing3b}), ours addresses the distant token problem by introducing the EABlock, which results in performance improvements.

\noindent \textbf{EABlock vs. previous Transformer blocks.}
Table~\ref{tab:rdtdesign} and Figure~\ref{fig:qual_abl} demonstrate the superiority of the EABlock over previous Transformer blocks.
For the first and second rows of Table~\ref{tab:rdtdesign}, we replaced our EABlock with standard Transformers with a similar number of parameters as EABlock, each with corresponding SA and CA modules.
Following the previous Transformer-based methods~\cite{li2022interacting,hampali2022keypoint}, we used two hand features as the input tokens.
The SA Transformer (the second row) extracts query, key, and value from both hands, and CA Transformer (the first row) extracts query from one hand and key-value pair from the other hand.
As shown, SA Transformer shows better results than CA Transformer, especially in the single hand cases.
This is because the CA mechanism forcefully injects irrelevant information from non-existing counterpart hand.

Meanwhile, the third row shows that our EABlock outperforms the two Transformer baselines only with the extract stage (\ie, without the adaptation stage) in EABlock.
Especially, the gain in two hands cases~(+1.68) is larger than that of single hand cases~(+0.78) in terms of MPJPE.
This demonstrates the benefits of our two novel tokens, which are proposed to relieve the distant token problem.
With the adaptation stage, our EABlock improves its performance in all metrics, shown in the fourth row.
In particular, the adaptation has more gain in single hand cases compared to two hand cases, both for MPJPE and MPVPE.
This justifies our adaptation stage, designed to adapt the interaction feature to each hand and filter irrelevant information from the interaction feature.

\begin{table}[t]
\def\arraystretch{1.2}
    \centering
    \resizebox{0.9\linewidth}{!}{
    \begin{tabular}{cc|ccc|ccc|c}
        \Xhline{2\arrayrulewidth}
         \multirow{2}{*}{$\mathbf{q}_\text{CA}$} & \multirow{2}{*}{$\mathbf{k}_\text{CA},\mathbf{v}_\text{CA}$} & \multicolumn{3}{c|}{MPJPE} & \multicolumn{3}{c|}{MPVPE} & \multirow{2}{*}{MRRPE}\\[-0.2em]
        & & Single & Two & All & Single & Two & All & \\
        \hline
        - & - & 13.19 & 14.17 & 13.88 & 9.68 & 13.61 & 11.95 & 25.95\\
        $\mathbf{t}_{\text{J}}$ & $\mathbf{F}_{\text{L}},\mathbf{F}_{\text{R}}$ & 10.66 & 12.60 & 11.63 & 8.73 & 11.65 & 10.00 & 21.88\\
        $\mathbf{t}_{\text{J}}$ & $\mathbf{t}_{\text{J}}$ & 12.02 & 13.42 & 12.71 & 8.79 & 11.63 & 10.02 & 22.67\\
        $\mathbf{t}_{\text{S}}$ & $\mathbf{t}_{\text{S}}$ & 10.44 & 12.36 & 11.40 & 8.53 & 11.33 & 9.75 & 21.70\\
        $\mathbf{t}_{\text{S}}$ & $\mathbf{t}_{\text{J}}$ & 10.48 & 12.40 & 11.44 & 8.58 & 11.37 & 9.79 & 21.52\\
        $\mathbf{t}_{\text{J}}$ & $\mathbf{t}_{\text{S}}$ & \textbf{9.62} & \textbf{11.54} & \textbf{10.58} & \textbf{7.86} & \textbf{10.78} & \textbf{9.13} & \textbf{18.82}\\
        \Xhline{2\arrayrulewidth}
    \end{tabular}}
    \vspace{-2mm}
    \caption{\textbf{Comparisons of models with various inputs for CA Transformer in FuseFormer.} 
    The first row uses $\mathbf{t}_{\text{J}}$ as the interaction feature without using the CA Transformer.}
    \vspace{-3mm}
    \label{tab:rdtin}
\end{table}

\begin{table}[t]
\def\arraystretch{1.2}
    \centering
    \resizebox{0.9\linewidth}{!}{
    \begin{tabular}{c|c|ccc|ccc|c}
        \Xhline{2\arrayrulewidth}
        \multirow{2}{*}{EABlock} & \multirow{2}{*}{SJT} & \multicolumn{3}{c|}{MPJPE} & \multicolumn{3}{c|}{MPVPE} & \multirow{2}{*}{MRRPE}\\[-0.2em]
        & & Single & Two & All & Single & Two & All & \\
        \hline
        \xmark & \xmark & 11.22 & 13.18 & 12.20 & 10.08 & 12.87 & 11.28 & 23.04\\
        \xmark & \cmark & 11.05 & 13.04 & 12.04 & 9.07 & 11.92 & 10.28 & 22.80\\
        \cmark & \xmark & 10.12 & 11.92 & 11.02 &  8.61 & 11.42 & 9.83 & 21.11\\
        \cmark & \cmark & \textbf{9.62} & \textbf{11.54} & \textbf{10.58} & \textbf{7.86} & \textbf{10.78} & 
        \textbf{9.13} & \textbf{18.82}\\
        \Xhline{2\arrayrulewidth}
    \end{tabular}}
    \vspace{-2mm}
    \caption{\textbf{Comparison of models with various combinations of EABlock and SJT.}}
    \vspace{-5mm}
    \label{tab:stgdesign}
\end{table}

\noindent\textbf{Inputs of CA Transformer in FuseFormer.}
Table~\ref{tab:rdtin} justifies our strategy to use JoinToken~$\mathbf{t}_{\text{J}}$ and SimToken~$\mathbf{t}_{\text{S}}$ as query and key-value pair of the CA Transformer in FuseFormer, respectively.
The comparison between the fifth row and our setting (the last row) shows that using JoinToken as a query produces lower 3D errors compared to using SimToken as a query.
This shows our setting effectively addresses the \emph{mismatch between query-key correlation and value}, described in Section~\ref{sec:eablock}.
In addition, our setting performs better than using a single type of token (the third and fourth rows), which indicates that our two tokens are complementary.
Ours also produces lower 3D errors compared to the second row that uses two hand features ($\mathbf{F}_{\text{L}}$ and $\mathbf{F}_{\text{R}}$) as a key-value pair.
This proves the benefit of using SimToken compared to the two hand features.
Lastly, ours produces far better results than the first row that uses JoinToken as the interaction feature without using the CA Transformer in FuseFormer.
This demonstrates that solely using JoinToken is insufficient, and the CA Transformer is necessary to fuse two types of input features of FuseFormer.

\begin{table}[t]
\def\arraystretch{1.2}
    \centering
    \resizebox{0.9\linewidth}{!}{
    \begin{tabular}{c|c|ccc|c}
        \Xhline{2\arrayrulewidth}
         \multirow{2}{*}{Methods} & \multirow{2}{*}{\# of params~(M)} & \multicolumn{3}{c|}{MPVPE}  & \multirow{2}{*}{MRRPE}\\[-0.2em]
        & & Single & Two & All & \\
        \hline
        Zhang~\etal~\cite{zhang2021interacting} & 143.37 & - & 13.95 & - & -\\
        Keypoint Transformer~\cite{hampali2022keypoint} & 117.05 & 10.16 & 14.36 & 11.94 & 30.87\\
        IntagHand~\cite{li2022interacting} & 39.04 & - & 9.03 & - & -\\
        \textbf{EANet~(Light. Ours)} & \textbf{33.90} & 5.71 & 7.66 & 6.53 & 34.29 \\
        \textbf{EANet~(Ours)} & 106.82 & \textbf{4.81} & \textbf{6.34} & \textbf{5.45} & \textbf{28.54} \\
        
        \Xhline{2\arrayrulewidth}
        
    \end{tabular}}
    \vspace{-2mm}
    \caption{\textbf{Quantitative comparison with state-of-the-art methods on InterHand2.6M~\cite{moon2020interhand2} dataset.}}
    \vspace{-3mm}
    \label{tab:sota25d_mesh}
\end{table}
\begin{table}[t]
\def\arraystretch{1.2}
    \centering
    \resizebox{0.9\linewidth}{!}{
    \begin{tabular}{c|ccc|c}
        \Xhline{2\arrayrulewidth}
         \multirow{2}{*}{Methods} & \multicolumn{3}{c|}{MPVPE}  & \multirow{2}{*}{MRRPE}\\[-0.2em]
        & Single & Two & All & \\
        \hline
        Zhang~\etal~\cite{zhang2021interacting} & - & 42.08 & - & -	\\
        Keypoint Transformer~\cite{hampali2022keypoint} & 60.19 & 51.21 & 54.71 & 190.77\\
        IntagHand~\cite{li2022interacting}  & - & 45.74 & - & - \\
        \textbf{EANet~(Light. Ours)} & 43.90 & 38.47 & 40.59 & 83.44 \\
        \textbf{EANet~(Ours)} & \textbf{32.82} & \textbf{34.43} & \textbf{33.80} & \textbf{81.11}\\
        \Xhline{2\arrayrulewidth}
    \end{tabular}}
    \vspace{-2mm}
    \caption{\textbf{Quantitative comparison with state-of-the-art methods on HIC~\cite{tzionas2016capturing} dataset.}
    }
    \vspace{-5mm}
    \label{tab:hic}
\end{table}

\noindent \textbf{Combinations of EABlock and SJT.}
Table~\ref{tab:stgdesign} shows that our SJT greatly harmonizes with the proposed EABlock.
Solely employing the SJT without EABlock, in the second row, shows a minor improvement.
Adopting EABlock alone, in the third row, shows considerable improvement compared to the baseline.
Most importantly, jointly adopting both EABlock and SJT further consistently improves the performance in every metric.

\begin{figure}[t]
\begin{center}
\captionsetup{justification=centering}
\captionsetup[sub]{font=scriptsize}
\begin{minipage}{1.\linewidth}
\centering
\includegraphics[width=1.0\linewidth]{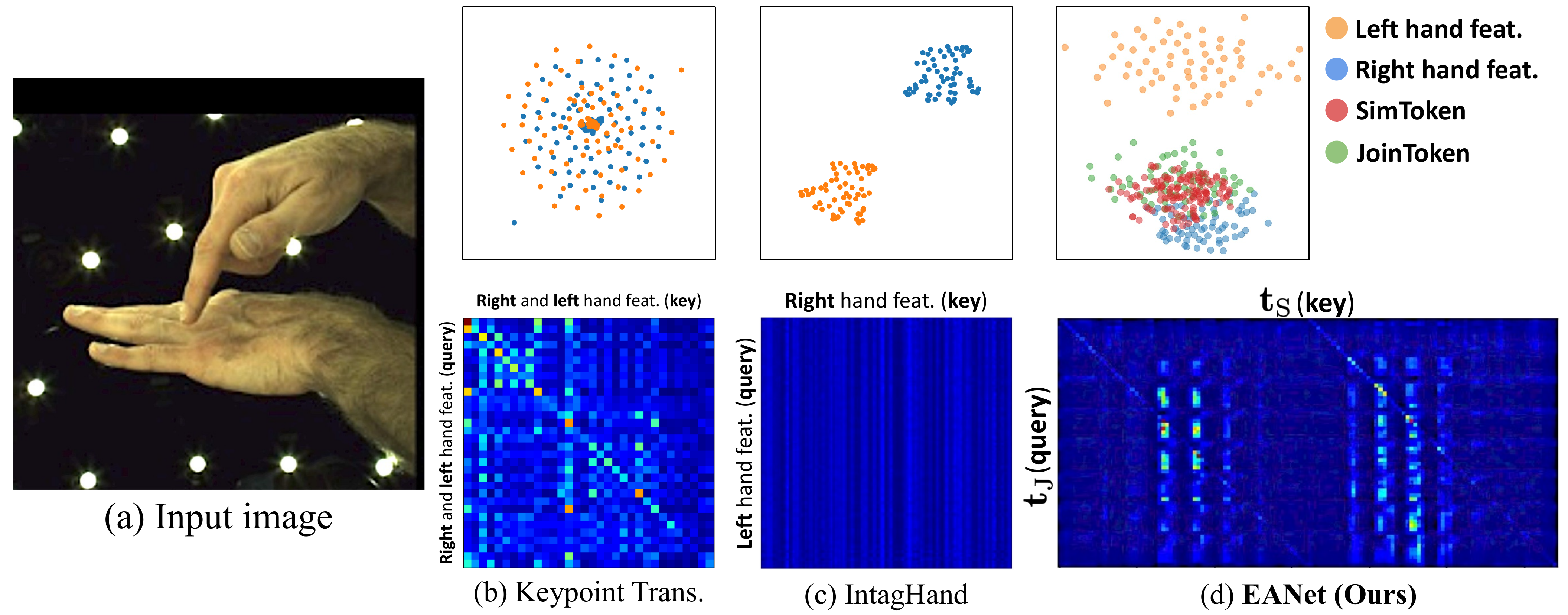}
\vspace{-4mm}
{\phantomsubcaption\label{fig:attn_mapa}
\phantomsubcaption\label{fig:attn_mapb}
\phantomsubcaption\label{fig:attn_mapc}
\phantomsubcaption\label{fig:attn_mapd}
}
\end{minipage}
\centering
\end{center}
\vspace{-7mm}
\caption{\textbf{Comparison of t-SNEs and attention maps.} 
Ours is from CA Transformer in extract stage of EABlock.
}
\vspace{-6mm}
\label{fig:attn_map}
\end{figure}

\subsection{Comparisons with state-of-the-art methods}

\noindent\textbf{Comparison on InterHand2.6M and HIC.}
Table~\ref{tab:sota25d_mesh} and \ref{tab:hic} show that our EANet achieves the highest performance on Interhand2.6M~\cite{moon2020interhand2} and HIC~\cite{tzionas2016capturing} datasets, respectively.
Following the previous works~\cite{li2022interacting,zhang2021interacting}, we measured MPVPEs after scale alignment on meshes with GTs.
Due to the absence of MPVPE in the manuscript of Keypoint Transformer, the numbers were obtained from their officially released codes and pre-trained weights.
In Table~\ref{tab:sota25d_mesh}, we compare the performance and the number of parameters with previous 3D interacting hand mesh recovering methods~\cite{moon2020interhand2, moon2022accurate, fan2021learning, hampali2022keypoint}.
Our EANet has fewer parameters than Keypoint Transformer~\cite{hampali2022keypoint} and Zhang~\etal~\cite{zhang2021interacting}, while significantly outperforming them.
Moreover, to ensure a fair comparison with IntagHand~\cite{li2022interacting}, which has fewer parameters than ours, we introduce a lighter version of EANet, EANet~(Light).
We reduced the channel dimension to $c = C/8$ instead of $c = C/4$ for EANet~(Light).
Even with fewer parameters, EANet~(Light) shows better MPVPE than IntagHand~\cite{li2022interacting}, further demonstrating the efficacy of our proposed approach.
Furthermore, in Table~\ref{tab:hic}, our EANet achieves significantly better performance on HIC dataset than other methods, indicating its strong generalizability.

In addition, Figure~\ref{fig:qualitative} shows the visual comparisons with previous state-of-the-art methods~\cite{li2022interacting, hampali2022keypoint} on InterHand2.6M~\cite{moon2020interhand2} and in-the-wild interacting hand images.
The results showcase our EANet's superior performance in precise hand pose and interaction estimation.

\noindent \textbf{Attention map comparison.}
Figure~\ref{fig:attn_map} compares attention maps and corresponding t-SNEs.
In Figure~\ref{fig:attn_mapb}, Keypoint Transformer~\cite{hampali2022keypoint} does not suffer from the distant token problem, as it does not separate feature from the backbone to left and right hand features.
However, without the separation, networks suffer from sub-optimal performance, as shown in Table~\ref{tab:projdesign}.
Also, its attention map shows the dominance of strong self-correlations with weak correlations from non-diagonal areas.
In Figure~\ref{fig:attn_mapc}, IntagHand suffers from the distant token problem.
Hence, it fails to generate proper correlations between two hands with different poses, resulting in predominantly low correlations between all queries and keys.
In contrast, Figure~\ref{fig:attn_mapd} shows that our SimToken and JoinToken do not suffer from the distant token problem with balanced high correlations in both the left- and right-half of the attention map.
As the each half of the attention map was computed between each query and each hand portion of SimToken, the high correlations in both halves indicate that ours effectively utilizes both left and right hand information for each query.

\noindent \textbf{Robustness to asymmetric poses of two hands.}
\begin{figure}[t]
    \centering
    \renewcommand{\wp}{0.52\linewidth}
    \includegraphics[width=\wp]{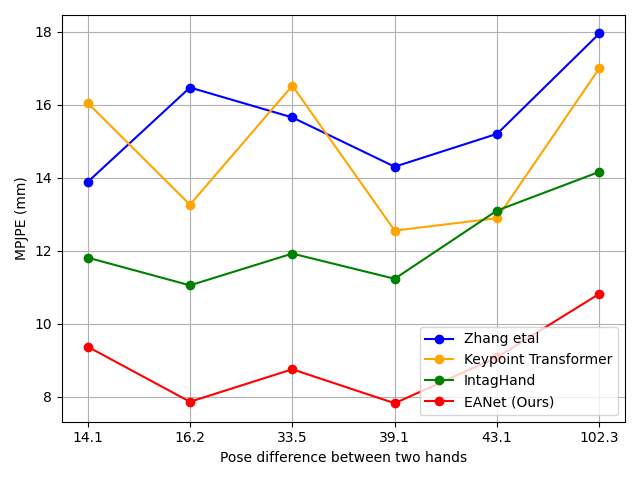}
    \vspace{-4mm}
    \caption{\textbf{Comparison on  InterHand2.6M~\cite{moon2020interhand2} sequences with various pose differences.} 
    $x$-axis represents pose difference between two hands.
    $y$-axis represents errors of the methods on each sequence with pose difference of $x$-axis.
    }
    \label{fig:splitcase}
    \vspace{-5mm}
\end{figure}
Figure~\ref{fig:splitcase} demonstrates that our EANet exhibits significantly greater robustness to asymmetric poses of the hands compared to previous state-of-the-art methods~\cite{zhang2021interacting,hampali2022keypoint,li2022interacting}.
For this analysis, we selected 6 sequences from the InterHand2.6M~\cite{moon2020interhand2}, which have diverse pose similarities between the two hands.
The $x$-axis of the figure represents the average pose difference between the two hands in each selected sequence, with larger values indicating bigger pose differences.
Notably, our EANet produces nearly consistent MPJPE values across all pose differences, even when the pose difference increases about 7 times at the rightmost $x$ value compared to the leftmost one.
Moreover, our method produces significantly lower MPJPE values than previous methods for all sequences, including those with the biggest pose differences.
We clarify the way to calculate the pose difference of each sequence in the supplementary material.

\section{Conclusion}
We present EANet, extract-and-adaptation network for 3D interacting hand mesh recovery.
With the help of the main block, EABlock, EANet effectively learns interaction between two hands by solving distant token problem.
The EABlock is mainly driven by our novel Transformer-based block, FuseFormer, which fuses two types of input features by using SimToken and JoinToken.
By extracting interaction from two hand features and then adapting the interaction towards each hand in EABlock, our EANet achieves the state-of-the-art performance in recovering 3D interacting hand mesh on challenging scenarios.

\noindent\textbf{Acknowledgments.}
This work was supported in part by the IITP grants [No.2021-0-01343, Artificial Intelligence Graduate School Program~(Seoul National University), No. 2021-0-02068, and No.2023-0-00156] and the NRF grant [No. 2021M3A9E4080782] funded by the Korea government~(MSIT).

\clearpage
\begin{center}
\textbf{\large Supplementary Material for \\ ``Extract-and-Adaptation Network for 3D Interacting Hand Mesh Recovery"}
\end{center}

\setcounter{section}{0}
\setcounter{table}{0}
\setcounter{figure}{0}

\renewcommand{\thesection}{\Alph{section}}   
\renewcommand{\thetable}{\Alph{table}}   
\renewcommand{\thefigure}{\Alph{figure}}

In this supplementary material, we provide more experiments, discussions, and other details that could not be included in the main text due to the lack of pages.
The contents are summarized below:

\begin{enumerate}[nosep, label=\Alph*., leftmargin=0.55cm]  
    \item Detailed architecture
    \begin{enumerate}[nosep, label=A\arabic*., leftmargin=0.6cm] 
        \item FuseFormer
        \item EABlock
    \end{enumerate}
    \item Number of adaptation stages in EABlock
    \item MPJPE comparison
    \item Measuring the pose difference
    \item Additional qualitative comparisons
    \item Discussion
    \begin{enumerate}[nosep, label=F\arabic*., leftmargin=0.6cm] 
        \item Limitations
        \item Societal impacts
    \end{enumerate}
\end{enumerate}

\algdef{SE}[SUBALG]{Indent}{EndIndent}{}{\algorithmicend\ }
\algtext*{Indent}
\algtext*{EndIndent}
\definecolor{commentcolor}{RGB}{110,154,155}   
\newcommand{\PyComment}[1]{\textcolor{commentcolor}{\# #1}}
\newcommand{\PyCode}[1]{\ttfamily\textcolor{orange}{\# #1}} 

\begin{algorithm}[!ht]
\caption{Pseudocode of FuseFormer in a PyTorch-style}
\label{algo:fuseformer}
	\begin{algorithmic}[1]
	    \State \textbf{class} FuseFormer(nn.Module): 
	        \Indent
	        \State def \textunderscore \textunderscore init \textunderscore\textunderscore():
	            \Indent
	            \State $\text{FC}$ = nn.Linear(1024, 512)
	            \State $\mathbf{t}_{\text{cls}}$ = nn.Parameter(512, 1)
	            \State Q\textunderscore SA = nn.Linear(512, 512)
	            \State K\textunderscore SA = nn.Linear(512, 512)
	            \State V\textunderscore SA = nn.Linear(512, 512)
	            \State Q\textunderscore CA = nn.Linear(512, 512)
	            \State K\textunderscore CA = nn.Linear(512, 512)
	            \State V\textunderscore CA = nn.Linear(512, 512)
	            \State $\psi$ : $\mathbb{R}^{1024\times 8\times 8}$ $\xrightarrow[]{}$ $\mathbb{R}^{64\times 1024}$  \PyComment{reshape}
	            \State $\phi$ : $\mathbb{R}^{512\times 8\times 8}$ $\xrightarrow[]{}$ $\mathbb{R}^{512\times 64}$  \PyComment{reshape}
	            \State $\eta$ : $\mathbb{R}^{512\times 64}$ $\xrightarrow[]{}$ $\mathbb{R}^{129\times 512}$  \PyComment{reshape}
	            \State $\text{softmax}$ = nn.Softmax(dim=-1)
	            \State $\text{MLP}$ = nn.Sequential(\\
	            {\color{white}aaaaaaaaaaaaaaaaaaaa} nn.Linear(512, 512*4),\\
	            {\color{white}aaaaaaaaaaaaaaaaaaaa} nn.Linear(512*4, 512))
	            \State $\text{MLP}'$ = nn.Sequential(\\
	            {\color{white}aaaaaaaaaaaaaaaaaaaa} nn.Linear(512, 512*4),\\
	            {\color{white}aaaaaaaaaaaaaaaaaaaa} nn.Linear(512*4, 512))
	            \State $d_{\mathbf{k}_{\text{SA}}}, d_{\mathbf{k}_{\text{CA}}} = 512, 512$
	            \EndIndent
	        \EndIndent
	        
	        \Indent
	        \State def forward($\mathbf{F}_{\text{L}}$, $\mathbf{F}_{\text{R}}$): \PyComment{$\mathbf{F}_{\text{L}} \text{ and } \mathbf{F}_{\text{R}} \in \mathbb{R}^{512\times 8\times 8}$}
    	        \Indent
    	        \State \PyComment{get JoinToken}
    	        \State $\mathbf{t}_{\text{J}}$ = $\text{FC}$($\psi$(torch.cat(($\mathbf{F}_{\text{L}}$,$\mathbf{F}_{\text{R}}$)))) \PyComment{$\in \mathbb{R}^{64 \times 512}$}
	            
	            \State \PyComment{get SimToken}
    	        \State $\mathbf{F}_{\text{L}}$ = $\phi(\mathbf{F}_{\text{L}})$\PyComment{$\in \mathbb{R}^{512 \times 64}$}
    	        \State $\mathbf{F}_{\text{R}}$ = $\phi(\mathbf{F}_{\text{R}})$\PyComment{$\in \mathbb{R}^{512 \times 64}$}
    	        \State $\mathbf{t}$ = $\eta$(torch.cat(($\mathbf{t}_{\text{cls}}$,$\mathbf{F}_{\text{L}}$,$\mathbf{F}_{\text{R}}$))) \PyComment{$\in \mathbb{R}^{129 \times 512}$}
    	        \State $\mathbf{q}_{\text{SA}}$, $\mathbf{k}_{\text{SA}}$, $\mathbf{v}_{\text{SA}}$ = Q\textunderscore SA($\mathbf{t}$), K\textunderscore SA($\mathbf{t}$), V\textunderscore SA($\mathbf{t}$)
    	        \State $\mathbf{r}$ = 
    	        $\text{softmax}$($(\mathbf{q}_{\text{SA}}@{\mathbf{k}_{\text{SA}}}^{T})/\sqrt{d_{\mathbf{k}_{\text{SA}}}}$)$\mathbf{v}_{\text{SA}}$ + $\mathbf{t}$
    	        \State $\mathbf{t}_{\text{S}}$ = $\mathbf{r}$ + $\text{MLP}$($\mathbf{r}$)
	            \State \PyComment{process two tokens}
	            \State $\mathbf{q}_{\text{CA}}$ = Q\textunderscore CA($\mathbf{t_{\text{J}}}$) 
    	        \State $\mathbf{k}_{\text{CA}}$, $\mathbf{v}_{\text{CA}}$ = K\textunderscore CA($\mathbf{t}_{\text{S}}$), V\textunderscore CA($\mathbf{t}_{\text{S}}$)
	            \State $\mathbf{r}'$ = 
    	        $\text{softmax}$($(\mathbf{q}_{\text{CA}}@{\mathbf{k}_{\text{CA}}}^{T})/\sqrt{d_{\mathbf{k}_{\text{CA}}}}$)$\mathbf{v}_{\text{CA}}$ + $\mathbf{t}_{\text{J}}$
    	        \State $\mathbf{F}_{\text{inter}}$ = $\mathbf{r}'$ + $\text{MLP}'$($\mathbf{r}'$)

	    \State \textbf{return} $\mathbf{F}_{\text{inter}}$
	        \EndIndent
	        \EndIndent
	\end{algorithmic}
\end{algorithm}

\section{Detailed architecture} \label{eablock_specific}
In the main manuscript, we introduce the Extract-and-Adaptation Network~(EANet) as a means of effectively extracting and processing interactions between two hands.
The proposed EANet mainly operates upon the key component, EABlock, to extract and adapt features from the two hands.
In this section, we provide a detailed explanation of the EABlock and its primary component, the FuseFormer, which are illustrated in Figures~\textcolor{red}{4} and \textcolor{red}{5} of the main manuscript, respectively.

\subsection{FuseFormer} \label{eablock_specific_adapt}
In Algorithm~\ref{algo:fuseformer}, we show the inference process of FuseFormer in the extract stage of EABlock.
From two hand features~($\mathbf{F}_{\text{L}}$ and $\mathbf{F}_{\text{R}}$), JoinToken~$\mathbf{t}_{\text{J}}$ is obtained by passing a concatenated two hand features to a fully connected layer after reshaping.
To obtain SimToken~$\mathbf{t}_{\text{S}}$, we first reshape each hand feature~($\mathbf{F}_{\text{L}}$ or $\mathbf{F}_{\text{R}}$) and concatenate the reshaped left and right hand features with a class token~$\mathbf{t}_{\text{cls}}$.
A concatenated token~$\mathbf{t}$ is made after reshaping the concatenated feature of $\mathbf{t}_{\text{cls}}$, $\mathbf{F}_{\text{L}}$, and $\mathbf{F}_{\text{R}}$.
The concatenated token~$\mathbf{t}$ is then used for extracting query~$\mathbf{q_\text{SA}}$, key~$\mathbf{k_\text{SA}}$, and value~$\mathbf{v_\text{SA}}$ with separate linear layers in a self-attention~(SA) based Transformer.
After preparing two tokens~($\mathbf{t}_{\text{J}}$ and $\mathbf{t}_{\text{S}}$), a cross-attention~(CA) based Transformer processes the two novel tokens, JoinToken~$\mathbf{t}_{\text{J}}$ as query~$\mathbf{q_\text{CA}}$ and SimToken~$\mathbf{t}_{\text{S}}$ as key-value pair~($\mathbf{k_\text{CA}}$ and $\mathbf{v_\text{CA}}$), to effectively fuse two hand features~($\mathbf{F}_{\text{L}}$ and $\mathbf{F}_{\text{R}}$) and output an interaction feature~$\mathbf{F}_{\text{inter}}$.

\subsection{EABlock} \label{eablock_specific_extract}
We show the inference process of EABlock in Algorithm~\ref{algo:EABlock}.
From left hand feature~$\mathbf{F}_{\text{L}}$ and right hand feature~$\mathbf{F}_{\text{R}}$, a FuseFormer in the extract stage of EABlock first extracts an interaction feature~$\mathbf{F}_{\text{inter}}$.
Then, with the extracted interaction feature~$\mathbf{F}_{\text{inter}}$ in the extract stage, each FuseFormer in the adaptation stage of EABlock fuses each hand feature~($\mathbf{F}_{\text{L}}$ or $\mathbf{F}_{\text{R}}$) and the interaction feature~$\mathbf{F}_{\text{inter}}$ to produce an adapted interaction feature for each hand~($\mathbf{F}_{\text{interL}}$ or $\mathbf{F}_{\text{interR}}$).
Lastly, our EABlock constructs the final outputs~($\mathbf{F}^{*}_{\text{L}}$ and $\mathbf{F}^{*}_{\text{R}}$) after a fully connected layer on each adapted interaction feature~($\mathbf{F}_{\text{interL}}$ or $\mathbf{F}_{\text{interR}}$) and then a concatenation to each hand feature~($\mathbf{F}_{\text{L}}$ or $\mathbf{F}_{\text{R}}$), respectively.

\algdef{SE}[SUBALG]{Indent}{EndIndent}{}{\algorithmicend\ }
\algtext*{Indent}
\algtext*{EndIndent}
\definecolor{commentcolor}{RGB}{110,154,155}   

\begin{algorithm}[!t]

\caption{Pseudocode of EABlock in a PyTorch-style}
\label{algo:EABlock}
	\begin{algorithmic}[1]
	    \State \textbf{class} EABlock(nn.Module): 
	        \Indent
	        \State def \textunderscore \textunderscore init \textunderscore\textunderscore():
	            \Indent
	            \State FuseFormer\textunderscore extract = FuseFormer()
	            \State FuseFormer\textunderscore L = FuseFormer()
	            \State FuseFormer\textunderscore R = FuseFormer()
	            \State FC = nn.Linear(512,128)
	            \EndIndent
	        \EndIndent
	        
	        \Indent
	        \State def forward($\mathbf{F}_{\text{L}}$, $\mathbf{F}_{\text{R}}$): \PyComment{$\mathbf{F}_{\text{L}} \text{ and } \mathbf{F}_{\text{R}} \in \mathbb{R}^{512\times 8\times 8}$}
    	        \Indent
    	        \State \PyComment{interaction feature extract}
    	        \State $\mathbf{F}_{\text{inter}}$ = FuseFormer\textunderscore extract($\mathbf{F}_{\text{L}}$, $\mathbf{F}_{\text{R}}$)
    	        \State \PyComment{interaction feature adaptation}
    	        \State $\mathbf{F}_{\text{interL}}$ = FuseFormer\textunderscore L($\mathbf{F}_{\text{L}}$, $\mathbf{F}_{\text{inter}}$)
    	        \State $\mathbf{F}_{\text{interR}}$ = FuseFormer\textunderscore R($\mathbf{F}_{\text{R}}$,  $\mathbf{F}_{\text{inter}}$)
    	        \State $\mathbf{F}^{*}_{\text{L}}$ = torch.cat(($\mathbf{F}_{\text{L}}$, FC($\mathbf{F}_{\text{interL}}$)))
    	         \State $\mathbf{F}^{*}_{\text{R}}$ = torch.cat(($\mathbf{F}_{\text{R}}$, FC($\mathbf{F}_{\text{interR}}$)))
	    \State \textbf{return}~$\mathbf{F}^{*}_{\text{L}}$, $\mathbf{F}^{*}_{\text{R}}$
	        \EndIndent
	        \EndIndent
	\end{algorithmic}
\end{algorithm}

\section{Number of adaptation stages in EABlock} \label{further_ablation}
To adapt the extracted interaction feature to each hand feature, our EABlock fuses an interaction feature and one hand feature~(\ie left hand feature or right hand feature) to obtain an adapted interaction feature for each hand.
Here, we justify our strategy to adapt only once to each hand feature.
Table~\ref{sup_tab:iter} shows that a single iteration of the adaptation stage shows comparable results to two iterations of the adaptation stages.
Considering the additional costs followed by additional adaptation stages, we set the number of adaptation stages to $1$.
\begin{table}[!h]
\def\arraystretch{1.2}
    \centering
    \resizebox{1\linewidth}{!}{
    \begin{tabular}{c|ccc|ccc|c}
        \Xhline{2\arrayrulewidth}
         \multirow{2}{*}{\# of adaptation} & \multicolumn{3}{c|}{MPJPE} & \multicolumn{3}{c|}{MPVPE} &  \multirow{2}{*}{MRRPE}\\
         & Single & Two & All & Single & Two & All & \\
        \hline
        0 & 10.66 & 11.81 & 11.25 & 8.37 & 11.24 & 9.61 & 21.00 \\
        \textbf{1}  & 9.62 & \textbf{11.54} & \textbf{10.58} & \textbf{7.86} & 10.78 & \textbf{9.13} & \textbf{18.82}\\
        2 & \textbf{9.60} & 11.90 & 10.75 & 8.14 & \textbf{10.66} & 9.19 & \textbf{18.82}\\
        
        \Xhline{2\arrayrulewidth}
    \end{tabular}}
    \vspace{-2mm}
    \caption{\textbf{Comparison of models with various number of adaptation stages.}
    }
    \vspace{-3mm}
    \label{sup_tab:iter}
\end{table}

\section{MPJPE comparison with state-of-the-art methods} \label{mpjpe}
Table~\textcolor{red}{5} in the main manuscript provides a comparison of MPVPE and MRRPE for various methods, while Table~\ref{sup_tab:sota_mpjpe} presents additional results based on MPJPE.
We follow the evaluation protocol of prior studies by using a pre-defined joint regression matrix to obtain the joint positions from the estimated meshes, which are scaled using the ground truth bone length.
Note that the MPJPE results for Keypoint Transformer in Table~\ref{sup_tab:sota_mpjpe} differ from those reported in their main manuscript, as our results are obtained using their official codes and pre-trained weights based on the mesh representation, while their reported numbers were based on a 2.5D representation.
Our results demonstrate that EANet outperforms other methods in 3D joint estimation, indicating its effectiveness.

\begin{table}[!h]
\def\arraystretch{1.2}
    \centering
    \resizebox{0.7\linewidth}{!}{
    \begin{tabular}{c|ccc}
        \Xhline{2\arrayrulewidth}
        \multirow{2}{*}{Methods}  & \multicolumn{3}{c}{MPJPE}\\
        &  Single & Two & All\\
        \hline
        Zhang~\etal~\cite{zhang2021interacting} & - & 13.48 & - \\
        Keypoint Transformer~\cite{hampali2022keypoint}  & 11.04 & 16.22 & 13.82 \\
        IntagHand~\cite{li2022interacting}  & - & 8.79 & - \\
        \textbf{EANet~(Ours)} &  \textbf{5.19} & \textbf{6.57} & \textbf{5.88} \\
        \Xhline{2\arrayrulewidth}
    \end{tabular}}
    \vspace{-2mm}
    \caption{\textbf{Quantitative comparison with state-of-the-art methods on InterHand2.6M~\cite{moon2020interhand2} dataset.}
    }
    \vspace{-3mm}
    \label{sup_tab:sota_mpjpe}
\end{table}

\section{Measuring the pose difference for Figure~\textcolor{red}{9}} \label{pose_difference}
In L833-L850 and Figure~\textcolor{red}{9}, we present the pose difference between two hands across different sequences of the InterHand2.6M dataset.
To compute the pose difference for each selected sequence, we begin by flipping a right hand to a left hand.
Next, we subtract the root joint~(wrist) position of each hand.
Finally, we align the global rotation of the two hands to solely determine the finger pose symmetricity.

\section{Additional qualitative comparisons}
\label{further_qual}
\begin{figure*}[t]
\begin{center}
    \vspace{0mm}
    \captionsetup{type=figure}
    \begin{center}
    \includegraphics[width=1\linewidth]{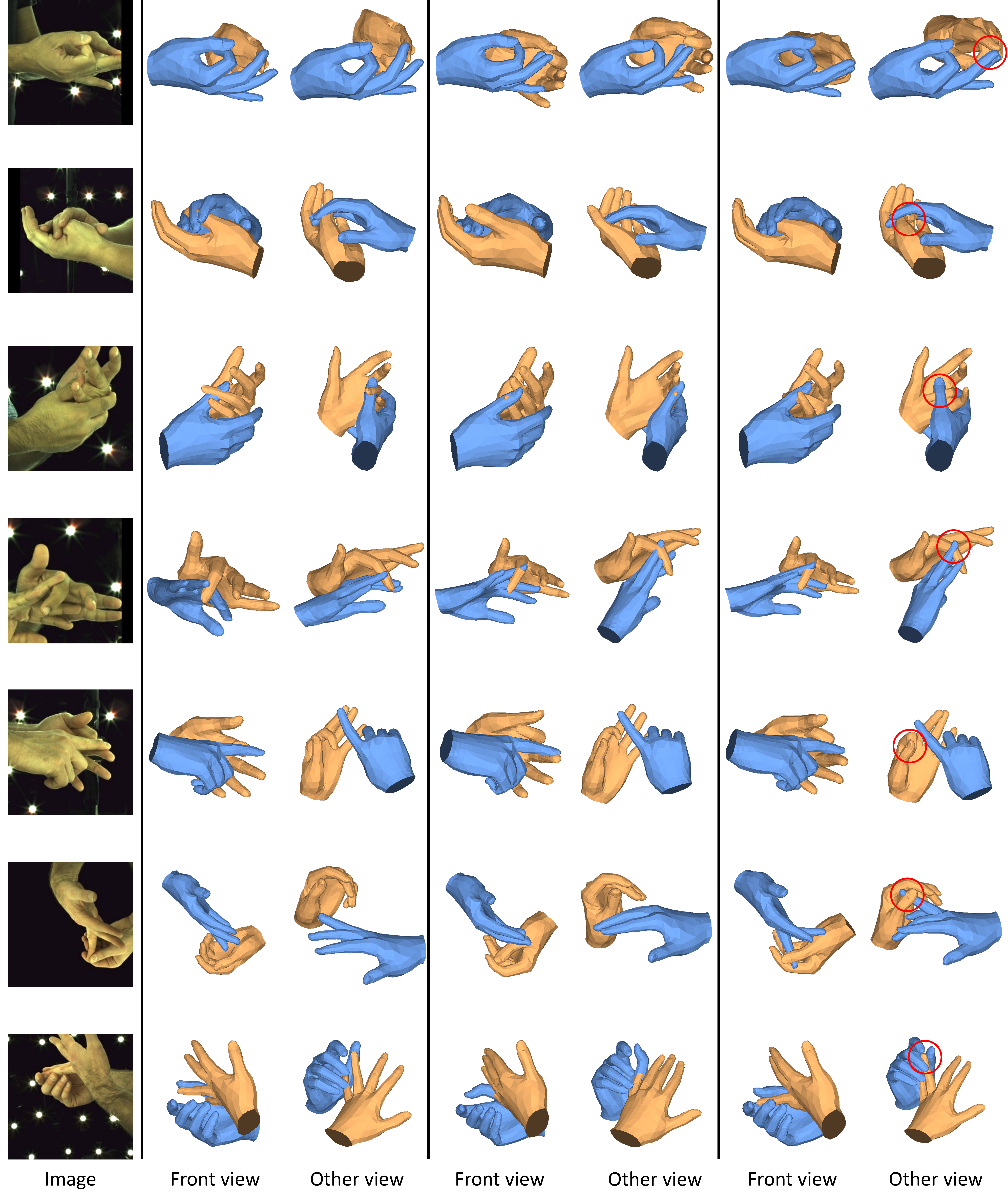}
    \end{center}
    \begin{minipage}{1.0\linewidth}
    \captionsetup[subfigure]{labelformat=empty}
    \subcaption{\textcolor{white}{AAAAAAAAAAAAAAAAAA}\textbf{IntagHand~\cite{li2022interacting}}\textcolor{white}{AAAAAAAAAA}\textbf{Keypoint Transformer~\cite{hampali2022keypoint}}\textcolor{white}{AAAAAAAAAA}\textbf{EANet~(Ours)}\textcolor{white}{AAAAAAAAA}}
    \end{minipage}
    \caption{\textbf{Visual comparison with state-of-the-art methods on InterHand2.6M~\cite{moon2020interhand2}. The red circles highlight regions where our EANet is correct, while others are wrong.}
    }
    \label{sup_fig:qual1}
    \vspace{-5mm}
\end{center}
\end{figure*}

\begin{figure*}[t]
\begin{center}
    \vspace{0mm}
    \captionsetup{type=figure}
    \begin{center}
    \includegraphics[width=1\linewidth]{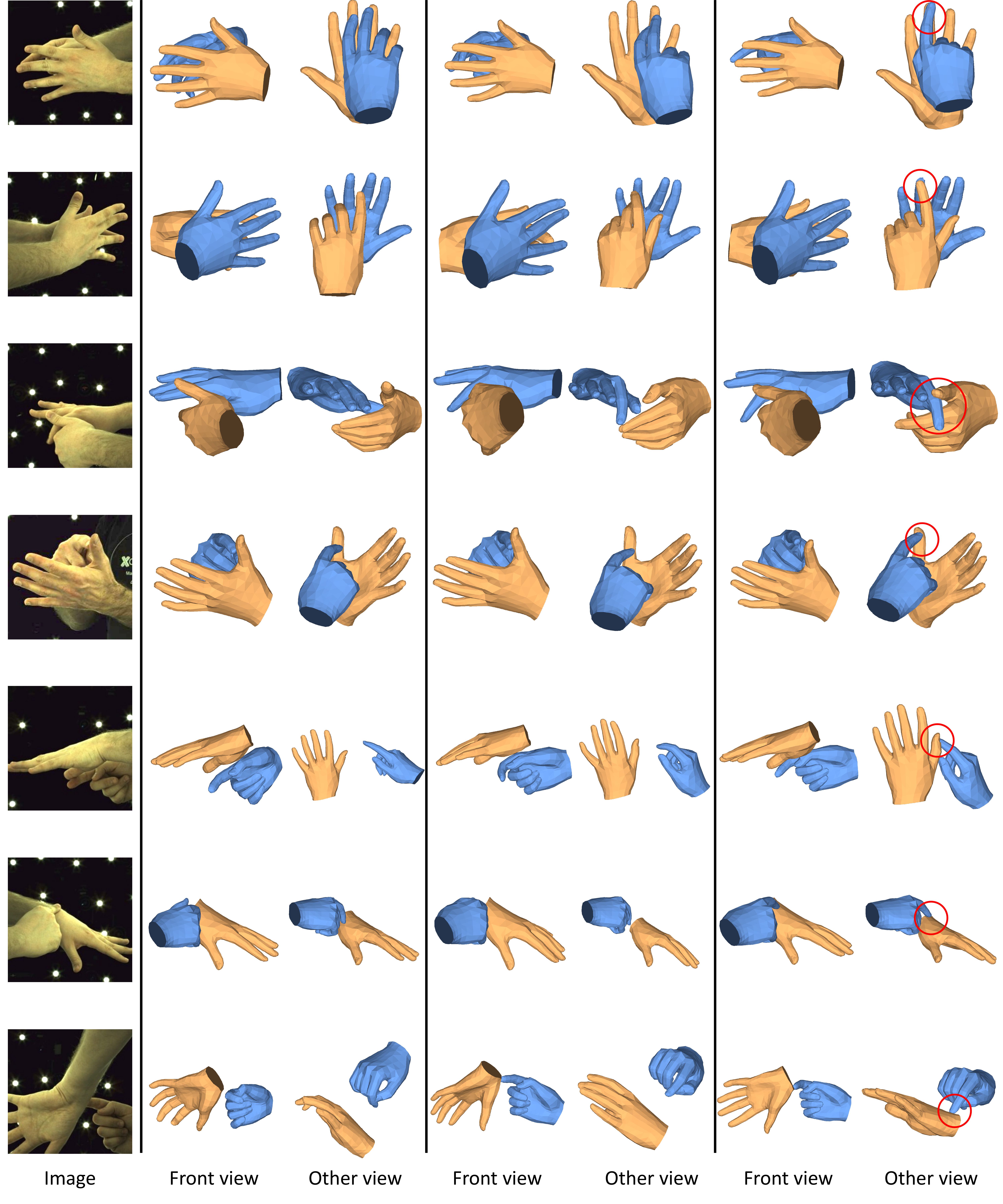}
    \end{center}
    \begin{minipage}{1.0\linewidth}
    \captionsetup[subfigure]{labelformat=empty}
    \subcaption{\textcolor{white}{AAAAAAAAAAAAAAAAAA}\textbf{IntagHand~\cite{li2022interacting}}\textcolor{white}{AAAAAAAAAA}\textbf{Keypoint Transformer~\cite{hampali2022keypoint}}\textcolor{white}{AAAAAAAAAA}\textbf{EANet~(Ours)}\textcolor{white}{AAAAAAAAA}}
    \end{minipage}
    \caption{\textbf{Visual comparison with state-of-the-art methods on InterHand2.6M~\cite{moon2020interhand2}. The red circles highlight regions where our EANet is correct, while others are wrong.}
    }
    \label{sup_fig:qual2}
    \vspace{-5mm}
\end{center}
\end{figure*}

In the below figures, we further show the qualitative comparisons with other state-of-the-art methods~\cite{li2022interacting,hampali2022keypoint}.
Specifically, in Figure~\ref{sup_fig:qual1} and \ref{sup_fig:qual2}, we show visual comparisons on InterHand2.6M~\cite{moon2020interhand2} dataset.
In Figure~\ref{sup_fig:qual3}, we show visual comparisons on in-the-wild interacting hand images.

\section{Discussion}
\subsection{Limitations}
\label{limitation}
\begin{figure*}[t]
\begin{center}
    \vspace{0mm}
    \captionsetup{type=figure}
    \begin{center}
    \includegraphics[width=1\linewidth]{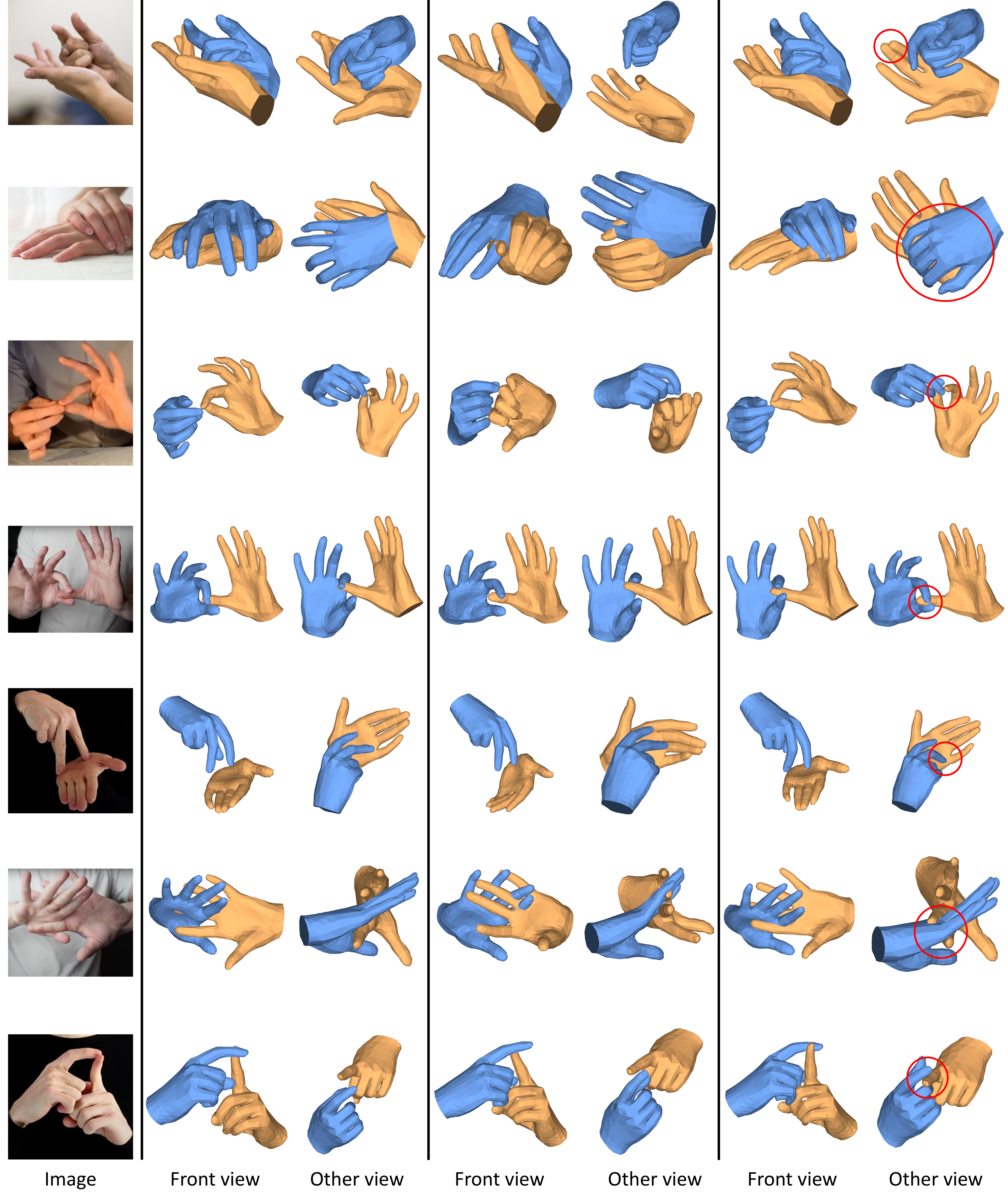}
    \end{center}
    \begin{minipage}{1.0\linewidth}
    \captionsetup[subfigure]{labelformat=empty}
    \subcaption{\textcolor{white}{AAAAAAAAAAAAAAAAAA}\textbf{IntagHand~\cite{li2022interacting}}\textcolor{white}{AAAAAAAAAA}\textbf{Keypoint Transformer~\cite{hampali2022keypoint}}\textcolor{white}{AAAAAAAAAA}\textbf{EANet~(Ours)}\textcolor{white}{AAAAAAAAA}}
    \end{minipage}
    \caption{\textbf{Visual comparison with state-of-the-art methods on in-the-wild images. The red circles highlight regions where our EANet is correct, while others are wrong.}
    The images are crawled from Google and YouTube.}
    \label{sup_fig:qual3}
    \vspace{-5mm}
\end{center}
\end{figure*}

Despite recent progress on domain adaptation~\cite{ganin2015unsupervised,pei2018multi,taigman2016unsupervised} that provides several techniques for reducing the discrepancy between training data and test data, our work has not yet employed such designs for better generalization ability.
In Figure~\ref{sup_fig:qual3}, our EANet shows reasonable results on in-the-wild images without such designs.
However, we believe that additional designs to improve generalization ability on in-the-wild images can further improve the performance of EANet in future works.

\subsection{Societal impacts}
\label{social_impact}
Our EANet performs robust 3D interacting hand mesh recovery with potential applications on technologies connecting real-world and virtual-world such as AR and VR.
Especially, the work is useful in cases where two hands are often interacting, including virtual meetings and virtual events.

\subsection*{License of the Used Assets}

\begin{compactitem}[$\bullet$]
    \item 
    \href{https://mks0601.github.io/InterHand2.6M/}{InterHand2.6M dataset~\cite{moon2020interhand2}} is CC-BY-NC 4.0 licensed. 
    \item \href{https://files.is.tue.mpg.de/dtzionas/Hand-Object-Capture/}{HIC dataset~\cite{tzionas2016capturing}} is publicly available dataset. 
    \item \href{https://github.com/BaowenZ/Two-Hand-Shape-Pose}{Zhang~\etal~\cite{zhang2021interacting} codes} are released for academic research only, and it is free to researchers from educational or research institutes for non-commercial purposes. 
    \item \href{https://github.com/shreyashampali/kypt_transformer}{Keypoint Transformer~\cite{hampali2022keypoint} codes} are released for academic research only, and it is free to researchers from educational or research institutes for non-commercial purposes. 
    \item \href{https://github.com/Dw1010/IntagHand}{IntagHand~\cite{li2022interacting} codes} are released for academic research only, and it is free to researchers from educational or research institutes for non-commercial purposes. 
\end{compactitem}

\clearpage

{\small
\bibliographystyle{ieee_fullname}

}

\end{document}